\documentclass[review]{elsarticle}
\usepackage{comment}
\usepackage{multirow}
\usepackage{graphicx}
\usepackage{color}
\usepackage{colortbl}
\usepackage{lineno}
\usepackage[colorlinks=true]{hyperref}
\usepackage{longtable}
\usepackage{subcaption}
\usepackage{rotating}
\usepackage{tikz} 
\usepackage{array} 
\usepackage{xfrac}
\usepackage[utf8]{inputenc}
\usepackage[T1]{fontenc}
\modulolinenumbers[5]
\usepackage[fleqn]{amsmath}
\usepackage{bm}
\usepackage{cancel}
\usepackage{booktabs}
\usepackage{ulem} 
\usepackage[utf8]{inputenc}
\usepackage[labelfont=bf]{caption}
\usepackage{threeparttable}
\usepackage{algorithm}
\usepackage{algpseudocode}
\usepackage{amsmath}
\usepackage{mathrsfs}
\usepackage{threeparttable}
\newcommand{\RNum}[1]{\uppercase\expandafter{\romannumeral #1\relax}}

\journal{arXiv.org}

\biboptions{authoryear}

\usepackage[left=2cm, right=2cm, top=2cm, bottom=2cm]{geometry}
\usepackage{amssymb}
\usepackage{amsmath}
\usepackage{latexsym}
\usepackage{color}
\usepackage{amsthm}
\usepackage[utf8]{inputenc}
\usepackage[english]{babel}

\usepackage{svg}
\usepackage[T1]{fontenc}
\usepackage{graphicx}

\let\originaleqref\eqref 
\makeatletter
\renewcommand{\eqref}[1]{%
  \begingroup%
  \let\ref\@refstar%
  \hyperref[#1]{%
    Eq.%
    ~\originaleqref{#1}%
18

  }%
  \endgroup
}
\makeatother

\begin{document}

\begin{frontmatter}

    \title{Modeling driver's evasive behavior during safety-critical lane changes: Two-dimensional time-to-collision and deep reinforcement learning}


\author[UC]{Hongyu Guo}

\author[OD]{Kun Xie\corref{mycorrespondingauthor}}

\author[UC]{Mehdi Keyvan-Ekbatani\corref{mycorrespondingauthor}}
\cortext[mycorrespondingauthor]{Corresponding author\\
\textit{E-mail addresses}: hongyu.guo@pg.canterbury.ac.nz (H.Guo),  
kxie@odu.edu (K.Xie),
mehdi.ekbatani@canterbury.ac.nz (M. Keyvan-Ekbatani)}

\address[UC]{Complex Transport Systems Laboratory (CTSLAB), Department of Civil and Natural Resources Engineering, University of Canterbury, Private Bag 4800, Christchurch 8140, New Zealand}
\address[OD]{Transportation Informatics Lab, Department of Civil and Environmental Engineering, Old Dominion University, 4635 Hampton Boulevard, Norfolk, VA 23529, United States}

\begin{abstract}
Lane changes are complex driving behaviors and frequently involve safety-critical situations. This study aims to develop a lane-change-related evasive behavior model, which can facilitate the development of safety-aware traffic simulations and predictive collision avoidance systems. Large-scale connected vehicle data from the Safety Pilot Model Deployment (SPMD) program were used for this study. A new surrogate safety measure, two-dimensional time-to-collision (2D-TTC), was proposed to identify the safety-critical situations during lane changes. The validity of 2D-TTC was confirmed by showing a high correlation between the detected conflict risks and the archived crashes. A deep deterministic policy gradient (DDPG) algorithm, which could learn the sequential decision-making process over continuous action spaces, was used to model the evasive behaviors in the identified safety-critical situations. The results showed the superiority of the proposed model in replicating both the longitudinal and lateral evasive behaviors.
\end{abstract}

\begin{keyword}
\texttt Evasive behavior \sep Lane change \sep Deep reinforcement learning \sep Surrogate safety measure \sep Big data analytics
\end{keyword}

\end{frontmatter}

\section{Introduction}

Lane changes are complicated driving behaviors that depend on the surrounding traffic dynamics, lane positions, and drivers' motivations \citep{sun2010modeling, keyvan2016categorization, ali2020cooperate, zhang2021spatiotemporal}. They might trigger conflicts in the interactions with the surrounding vehicles and activate the drivers' evasive actions \citep{zheng2010impact, zheng2014recent, chen2021modeling}. At the same time, lane changes could also be evasive actions in response to the potential risks \citep{petersen2009postural, xu2012dynamic}. It is critical to understand evasive behaviors during lane changes, given their complexity and significant safety impact. The accurate modeling of the vehicle's motions under safety-critical conditions could contribute to the safety-aware traffic simulation \citep{gettman2003surrogate, markkula2012review, wang2018combined}. Besides, the advanced driving assistance system (ADAS) and the automated vehicle can incorporate useful information from the evasive behavior model to provide driver-specific control intervention and plan an evasive path consistent with human's operating habits \citep{tseng2005evasive, happee2017take, soudbakhsh2011emergency}. 

The accurate detection and extraction of safety-critical driving scenarios are fundamental for studying evasive behaviors. Compared to the traditional methods, which rely on historical crashes that often require a significant time to collect sufficient data \citep{scanlon2015analysis}, the approaches based on surrogate safety measures (SSMs) are more proactive. The SSMs could capture the more frequent "near-crash" situations and help understand the process of evasions or crashes \citep{li2020analysis, li2021exploring}. Many studies have used SSMs for risk identification and their correlation with the crash data has been affirmed \citep{xie2016development, xie2019use, yang2021fusing}. It is well accepted that the SSM-based approach can be used to identify and analyze safety-critical scenarios. 

As an integral effort to improve traffic safety, driver's evasive behavior models have received wide attention \citep{markkula2012review, dozza2013factors, xiong2019forward, scanlon2021waymo}. Many kinematic-based approaches have been proposed to simulate the braking and/or steering behavior in near-crash situations \citep{markkula2016farewell, svard2021computational}. Meanwhile, the reinforcement learning (RL) methods, which can capture and model the drivers' actions sequentially, have been used in driving behavior modeling in recent years \citep{haydari2020deep, farazi2021deep}. The RL algorithms could be used to model both the upper-level decision making process \citep{li2022decision} and the lower-level vehicle motion control \citep{zhu2018human}. The safety risk was usually considered as a component of the reward function \citep{zhu2020safe, li2022decision}. Limited studies employed RL for evasive behavior modeling in pedestrian-related conflicts \citep{zuo2020microscopic, nasernejad2021modeling, papini2021reinforcement}. It is worth to further unlock the potential of RL approaches in driver's evasive behavior modeling.

Video recording and driving simulator data were widely used to study evasive behaviors. These datasets either only cover limited driving scenarios or do not reflect the real driving experience. With the onboard sensors, the connected vehicles (CVs) could collect large-scale naturalistic motion data of the subject vehicle and its ambient traffic, which is beneficial for driver behavior modeling \citep{zhao2017trafficnet, guo2021lane, guo2022lane}. The Safety Pilot Model Deployment (SPMD) program is a real-world CV test conducted by the U.S. Department of Transportation (USDOT) \citep{henclewood2014safety}. It provides rich high-resolution data for the identification of safety-critical situations and the development of evasive behavior models.

This study aims to develop a lane-change-related evasive behavior model using the CV data from the SPMD program. A novel SSM is proposed to capture the safety risk during lane changes. The deep deterministic policy gradient (DDPG) \citep{lillicrap2015continuous} algorithm is used to imitate the drivers' evasive behaviors. The remainder of this paper is organized as follows. In Section \ref{literature}, a thorough literature review on the related research is carried out. The proposed two-dimensional time-to-collision (2D-TTC) and the DDPG model are introduced in Section \ref{methodology}. The detailed data processing procedure is illustrated in Section \ref{data}. In Section \ref{results}, the results and discussions are presented. At last, the conclusions of this study are summarized in Section \ref{conclusion}.


\section{Literature review}\label{literature}

\subsection{Surrogate safety measures}

In the safety-related research, SSM is a proactive approach by capturing the more frequent near-crash situations, compared to the historical crash records. Time to collision (TTC) is one of the most widely used SSMs, which was introduced by \cite{hayward1972near}. It is defined as the time required for two vehicles to collide if they continue in their present speed along the same path, and is calculated as

\begin{equation}
\begin{aligned}
TTC=\left\{\begin{matrix}
\frac{s_{0}-l}{v-v_{0}}, &v > v_0\\
\infty, &otherwise
\end{matrix}\right.
\end{aligned}
\end{equation}

where $s_0$ is the space headway between the following and leading vehicles, $l$ is the length of the leading vehicle, $v_0$ and $v$ are the initial velocity of the leading and following vehicles, respectively. If the TTC value is less than a threshold, the car-following scenario is considered to be unsafe. Due to its simplicity and practicality, TTC has been widely used as a safety indicator in many studies \citep{lenard2018time, li2020analysis, deveaux2021extraction, li2021exploring}. Based on TTC, \cite{minderhoud2001extended} proposed time exposed TTC (TET) and time integrated TTC (TIT). They were able to include the duration of dangerous driving scenarios into risk measurement and were also applied in several studies \citep{jamson2013behavioural,li2017evaluation, yang2021proactive}.

However, there are two major shortcomings of the conventional TCC: i) the scenario is regarded as safe when the speed of the following vehicle is less than or equal to that of the leading vehicle, even though the relative distance could be very small \citep{kuang2015tree}; and ii) the vehicle pair is assumed in the same lane and only the longitudinal movements are calculated \citep{xing2019examining}. To address these limitations, some researchers included the speed changes of vehicles into TTC. \cite{ozbay2008derivation} modified TTC by considering all the potential longitudinal conflicts related to acceleration or deceleration. \cite{xie2019use} imposed a hypothetical disturbance to the leading vehicle and derived the TTC with disturbance (TTCD). It has been proved that TTCD was capable to identify high-risk locations and assess safety performance with real-world data \citep{yang2021fusing}. 

Some other studies concentrated on expanding the TTC to the two-dimensional road plane and proposed the trajectory-based SSMs. \cite{hou2014new} proposed algorithms for computing TTC by solving the two-dimensional distance equations and validated them with simulation models. \cite{ward2015extending} generalized TTC to the two-dimensional movement case and derived a method for predicting conflicting trajectories based on the relative movement between two vehicles. \cite{venthuruthiyil2022anticipated} proposed anticipated collision time (ACT) for the two-dimensional trajectory-based proactive safety assessment. However, these SSMs relied on the recorded complete trajectories and could not estimate the potential risks instantly. In this study, the CVs are probe vehicles, moving around and collecting data only from themselves and surrounding vehicles. It is expensive to reconstruct the complete trajectories of all vehicles from the CV data. Furthermore, to facilitate collision avoidance systems, safety risks should be captured in real time instead of offline. A new TTC that is customized for CV data and capable for capturing the conflict instantaneously is needed.

\subsection{Evasive behavior}

As one of the bases to improve traffic safety, it is crucial to understand and model the evasive behaviors of drivers accurately under safety-critical conditions \citep{markkula2012review}. Many researchers have been focusing on evasive behavior modeling, and the findings can be categorized into three groups according to the maneuvers, namely evading by braking (longitudinal acceleration) alone, steering (lateral acceleration) alone, and the interplay of braking and steering. 

Both the naturalistic and simulator-based studies indicate that braking is the most common avoidance response \citep{mcgehee1999examination, najm2013description}. Thus, the braking behavior in pre-crash situations has been a research hot spot \citep{markkula2016farewell, svard2017quantitative, xue2018using, svard2021computational}. The steering behavior is less common in evasion, as it might involve more decision factors, such as the surrounding traffic situations. It is expected to help avoid the collision that cannot be avoided by braking only and has drawn some research attention \citep{shibata2014collision, yuan2019unified, zheng2020bezier}. 

The combination of braking and steering evasive behaviors has the potential to further reduce the probability of collisions \citep{horiuchi2001numerical, schmidt2006research}. However, there are limited studies on the combined evasive behavior due to its complexity. Some studies focused on predicting the driver's choice of evasive behavior (braking or steering) using classification methods \citep{venkatraman2016steer, sarkar2021steering}. Further studies focused on modeling the driver's braking and steering actions. \cite{sugimoto2005effectiveness} assumed an evasive behavior model including braking and steering, and used it to evaluate the effectiveness of ADAS. \cite{jurecki2009driver} designed a model for collision avoidance at intersections based on the results of track tests. Both the barking and steering behaviors could be reconstructed successfully in pre-crash situations. \cite{markkula2014modeling} proposed a framework to model the drivers’ steering and pedal behaviors as a series of individual control adjustments. \cite{schnelle2018feedforward} presented an integrated feedback and feedforward driver model, which could represent the driver's control actions in safety-critical scenarios. \cite{zhou2020evasive} developed statistical models from near-crash trajectories and drew evasive behaviors from the joint probability distributions. 

In most studies, the evasive behavior models were built based on the data from driving simulators or track tests, which may not reflect real-world driving situations. A limited number of studies adopted the crash and near-crash records for model development \citep{markkula2016farewell, zhou2020evasive, svard2021computational, sarkar2021steering}, while the historical crashes data often requires a significant time for collection. Besides, the kinematic-model-based approaches were widely used in these studies. The models with assumed constraints might be not flexible enough to simulate evasive behavior, since the extreme driving behavior is likely to be taken in safety-critical situations. Meanwhile, the data-driven methods have been widely applied in driving behavior modeling due to their flexibility and outperformed accuracy \citep{huang2018car, lee2019integrated, xing2020ensemble, ma2020sequence}. It is worth to develop a proactive SSM-based approach for risk detection and build an evasive behavior model with data-driven approaches.

\subsection{Reinforcement learning in driving behavior modeling}

RL is a machine learning approach aiming to let an intelligent agent interact with an environment and make decisions. The agent is trained to learn the optimal policy that maximizes the reward function. The basic RL is modeled as a Markov decision process (MDP). At each time step $t$, the RL agent is at a state $S_t$. It chooses an action $A_t$ according to the policy $\pi(A_t|S_t)$, which is subsequently sent to the environment. Then, the agent moves to a new state $S_{t+1}$ and receives a reward $R_t$. The MDP continues until the system reaches a terminal state and it will restart. The agent aims to maximize the discounted accumulated reward function $\sum R_t = \sum_{k=0}^{\infty} \gamma^k R_{t+k}$, with the discount factor $\gamma \in (0,1]$ \citep{li2017deep}. As the RL algorithm improves its policies by interacting with the environment, it is suitable for problems whose solutions can be optimized by trial-and-error. It is also appropriate for the problems that emphasize task completion and delayed reward over the periodical success at intermediate steps. Because of these features, RL was widely applied in transportation studies and provided promising results \citep{abdulhai2003reinforcement, ozan2015modified, aslani2017adaptive, essa2020self}. 

Deep reinforcement learning (DRL) is the combination of RL and deep learning (DL). The policy and other learned functions are often represented as neural networks. As DL is incorporated into the RL solution, the agent is able to make decisions from complex and high-dimensional raw input data with less manual engineering of the state space. Due to the ability to generate high-quality solutions and generality in solving varying problems, DRL has been widely used in driving behavior modeling aiming at the application of autonomous vehicles \citep{haydari2020deep, farazi2021deep}. It could be used for both the upper-level driving behavior decision making \citep{dong2021space, wang2021harmonious, li2022decision} and the lower-level vehicle motion control \citep{zhu2018human, ye2019automated, zhu2020safe, tian2021learning}. 

As for the application of RL in traffic safety, the safety factors were often included and modeled as a component of the reward function in some of the aforementioned studies \citep{ye2019automated, zhu2020safe, dong2021space}. \cite{li2022decision} assessed the risk of lane change decision and used a deep Q-network (DQN), a widely used model of DRL, to find the strategy with the minimum expected risk. The model outperformed the baseline models without risk assessment in terms of average traveled distance. There was also limited research using DRL to model the drivers' evasive behaviors in critical situations. \cite{chae2017autonomous} developed an autonomous braking system with DQN, which decided whether to apply the brake when confronting the risk of collision. \cite{zuo2020microscopic} used RL to determine the optimal timing to cross the intersection with the consideration of pedestrian safety. \cite{papini2021reinforcement} studied the scenario that a pedestrian suddenly crossing the road and employed DQN to choose an appropriate speed, which enabled the autonomous vehicle to perform an emergency braking maneuver successfully. Besides, the RL approaches were used to model the behaviors in pedestrian-cyclist conflicts \citep{nasernejad2021modeling, alsaleh2021markov}.

Most of the existing studies are developed based on traffic simulations, as it is a noise-free interactive environment for the RL agent. Only a small portion of previously mentioned research used human driving data, either the driving simulator data \citep{tian2021learning} or the video-based real-world data \citep{zhu2020safe, zhu2018human, nasernejad2021modeling, alsaleh2021markov, shi2021connected}. Although the models achieved good performance in the simulation environment, they might not be able to reflect real situations. Additionally, the video-based data are collected with a limited range of roads or intersections, which may not cover various traffic scenarios under different conditions. It is critical to develop the evasive behavior model using large-scale real-world data. 

\section{Methodology}\label{methodology}
\subsection{2-dimensional time-to-collision}


To address the limitations of TTC discussed in the literature review, we extended the TTC to a two-dimensional plane and proposed a new TTC called 2D-TTC. The proposed approach relaxes the constraint of the vehicle pair's positions and motions. It is also suitable for real-time risk measurement based on onboard sensors of CVs.

Similar to TTC, the leading and following vehicles are assumed to maintain their current speeds and directions. The trajectory of each vehicle will be estimated, and the collision is detected based on their estimated positions. Both the longitudinal and lateral risks could be captured by 2D-TTC. Additionally, it is intuitive as it is based on the well-known and widely used concept of TTC.

\begin{figure}[!htpb]
\centering\includegraphics[width=0.5\linewidth]{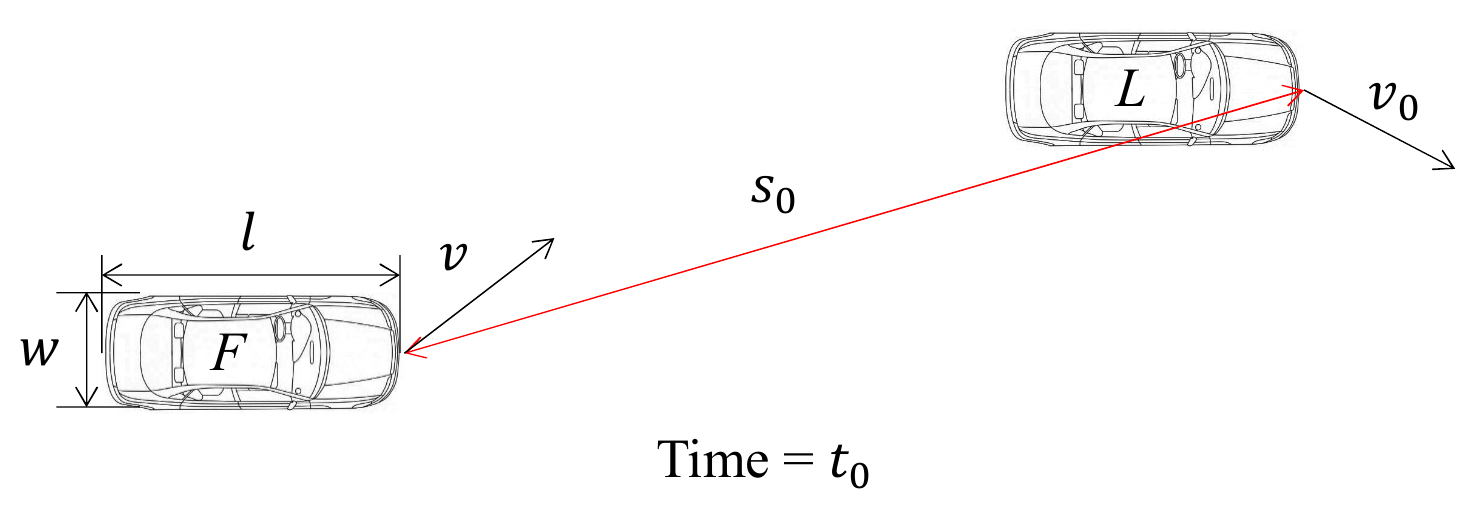}
\caption{A 2-D traffic scenario}\label{basic_scenario}
\end{figure}

A general two-dimensional traffic scenario at the initial time $t_{0}$ is illustrated in Fig. \ref{basic_scenario}. $s_{0}$ is the spacing headway between the two vehicles, $v_{0}$ and $v$ are the speed of the leading and following vehicles, respectively. It is assumed that the two vehicles have the same length $l$ and width $w$. The longitudinal and lateral components of $s_{0}$, $v_{0}$, and $v$ are denoted as $s_{0lon}$, $v_{0lon}$, $v_{lon}$, and $s_{0lat}$, $v_{0lat}$, $v_{lat}$, respectively.

Depending on the relative positions of the two vehicles, there are two possible collision outcomes: i) longitudinal collision - the front end of the following vehicle hit the rear end of the leading vehicle; and ii) lateral collision - one side of the following vehicle hit the other side of the leading vehicle. These two types of collision will be analyzed and computed separately.

\begin{itemize}
    \item \textit{Longitudinal collision and longitudinal time to collision}
\end{itemize}

The longitudinal time-to-collision ($TTC_{lon}$) is defined as the time required for the front end of the following vehicle to reach the same longitudinal position as the rear end of the leading vehicle. If $s_{0lon}>l$ and $v_{lon}>v_{0lon}$, $TTC_{lon}$ is calculated as

\begin{equation}
TTC_{lon}=\frac{s_{0lon}-l}{v_{lon}-v_{0lon}}
\end{equation}

\begin{figure}[!htpb]
\centering\includegraphics[width=0.7\linewidth]{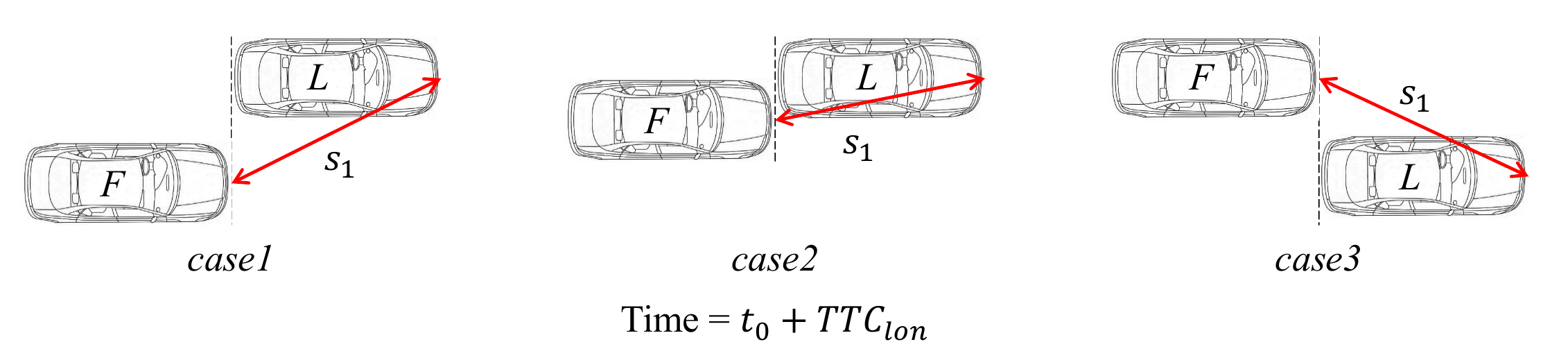}
\caption{Longitudinal collision scenarios}\label{rear_end}
\end{figure}

When the two vehicles arrive at the assumed positions at time $t_{0}+TTC_{lon}$, there are three possible cases of their relative positions, as shown in Fig. \ref{rear_end}. 

The occurrence of collision depends on the remaining lateral distance $s_{1lat}$, which is given by Eq. (\ref{s1lat}):

\begin{equation}\label{s1lat}
s_{1lat} = s_{0lat}-(v_{lat}-v_{0lat}) \cdot TTC_{lon} =  s_{0lat}-(v_{lat}-v_{0lat}) \cdot \frac{s_{0lon}-l}{v_{lon}-v_{0lon}}
\end{equation}

If $s_{1lat} > w$ (case1 or case3), there is no lateral overlap of the two vehicles, and thus no collision occurs. If $s_{1lat}\leq w$ (case2), a rear-end collision would happen. By summarizing the computation and collision conditions, $TTC_{lon}$ is calculated as 

\begin{equation}
\begin{aligned}
TTC_{lon}=\left\{\begin{matrix}
\frac{s_{0lon}-l}{v_{lon}-v_{0lon}}, &s_{0lon} > l \wedge v_{lon} > v_{0lon} \wedge s_{0lat}-(v_{lat}-v_{0lat}) \cdot \frac{s_{0lon}-l}{v_{lon}-v_{0lon}}<w\\
\infty, &otherwise
\end{matrix}\right.
\end{aligned}
\end{equation}

\begin{itemize}
    \item \textit{Lateral collision and lateral time-to-collision} 
\end{itemize}

The lateral time-to-collision ($TTC_{lat}$) is defined as the time required for the left (right) side of the following vehicle to reach the same lateral position as the right (left) side of the leading vehicle. If $s_{0lat} > w$ and $v_{lat}>v_{0lat}$, $TTC_{lat}$ is calculated as

\begin{equation}
TTC_{lat} = \frac{s_{0lat}-w}{v_{lat}-v_{0lat}}
\end{equation}

\begin{figure}[!htpb]
\centering\includegraphics[width=0.7\linewidth]{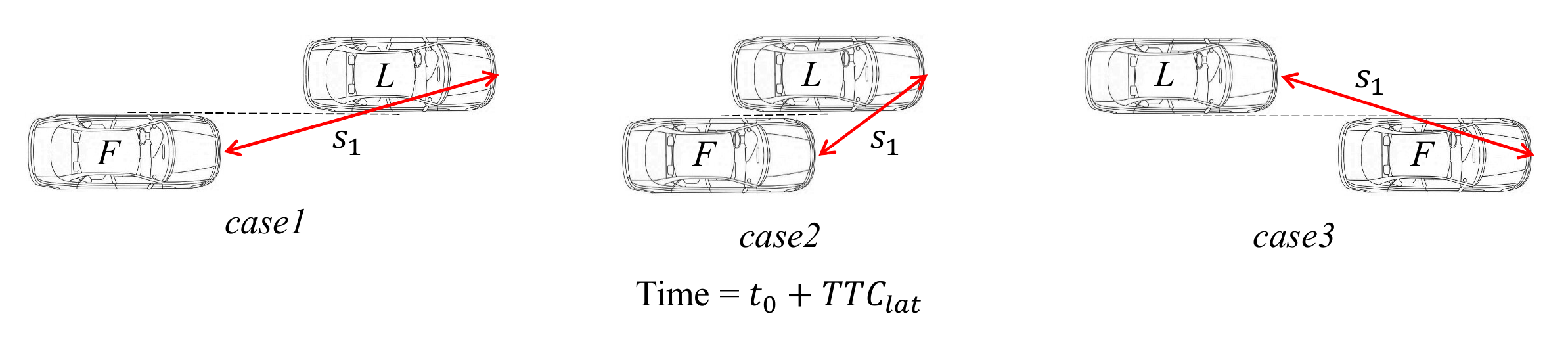}
\caption{Sideswipe collision scenario}\label{sideswipe}
\end{figure}

When the two vehicles arrive at the assumed positions at time $t_{0}+TTC_{lat}$, there are three possible cases of their relative positions, as shown in Fig. \ref{sideswipe}.

The occurrence of collision depends on the remaining longitudinal distance $s_{1lon}$, which is given by Eq. \ref{s1lon}.

\begin{equation}\label{s1lon}
s_{1lon} = s_{0lon}-(v_{lon}-v_{0lon}) \cdot TTC_{lat} =  s_{0lon}-(v_{lon}-v_{0lon}) \cdot \frac{s_{0lat}-w}{v_{lat}-v_{0lat}}
\end{equation}

If $s_{1lon} > l$ (case1 or case3), there is no longitudinal overlap of the two vehicles, and thus no collision occurs. If $s_{1lon} \leq l$ (case2), a sideswipe collision would happen. By summarizing the computation and collision conditions, $TTC_{lat}$ is calculated as 

\begin{equation}
\begin{aligned}
TTC_{lat}=\left\{\begin{matrix}
\frac{s_{0lat}-w}{v_{lat}-v_{0lat}}, &s_{0lat} > w \wedge v_{lat} > v_{0lat} \wedge s_{0lon}-(v_{lon}-v_{0lon}) \cdot \frac{s_{0lat}-w}{v_{lat}-v_{0lat}}<l\\
\infty, &otherwise
\end{matrix}\right.
\end{aligned}
\end{equation}

By combining the computations of $TTC_{lon}$ and $TTC_{lat}$, the unified function of $TTC_{2D}$ is expressed as

\begin{equation}
TTC_{2D}=min(TTC_{lon}, TTC_{lat})
\end{equation}

The type of conflict can also be classified with 2D-TTC. Specifically, if $TTC_{lon} \leq TTC_{lat}$, there would be a potential longitudinal (rear-end) crash; if $TTC_{lat}<TTC_{lon}$, there would be a potential lateral (sideswipe) crash. In the implementation of 2D-TTC, both the longitudinal and lateral conflicts should be considered jointly. Because the types of captured conflicts are interchangeable with the disturbances. Similar to TTC, an appropriate threshold is needed to differentiate the risky and safe scenarios. The vehicle pair with 2D-TTC lower than the threshold is involved in conflicts. 

Clearly, the 2D-TTC is not only related to the relative distance and speed of the two vehicles but also their initial positions and directions. In this case, the vehicle pair not in the same path (i.e. a cut-in vehicle from the adjacent lane) could still yield risk rather than be regarded as safe by the conventional TTC. Besides, the 2D-TTC is computed based on the predicted trajectories, which overcomes the shortcomings of trajectory-based SSMs and is capable for real-time risk estimation.

\subsection{Deep deterministic policy gradient}

\begin{figure}[!htpb]
\centering\includegraphics[width=0.7\linewidth]{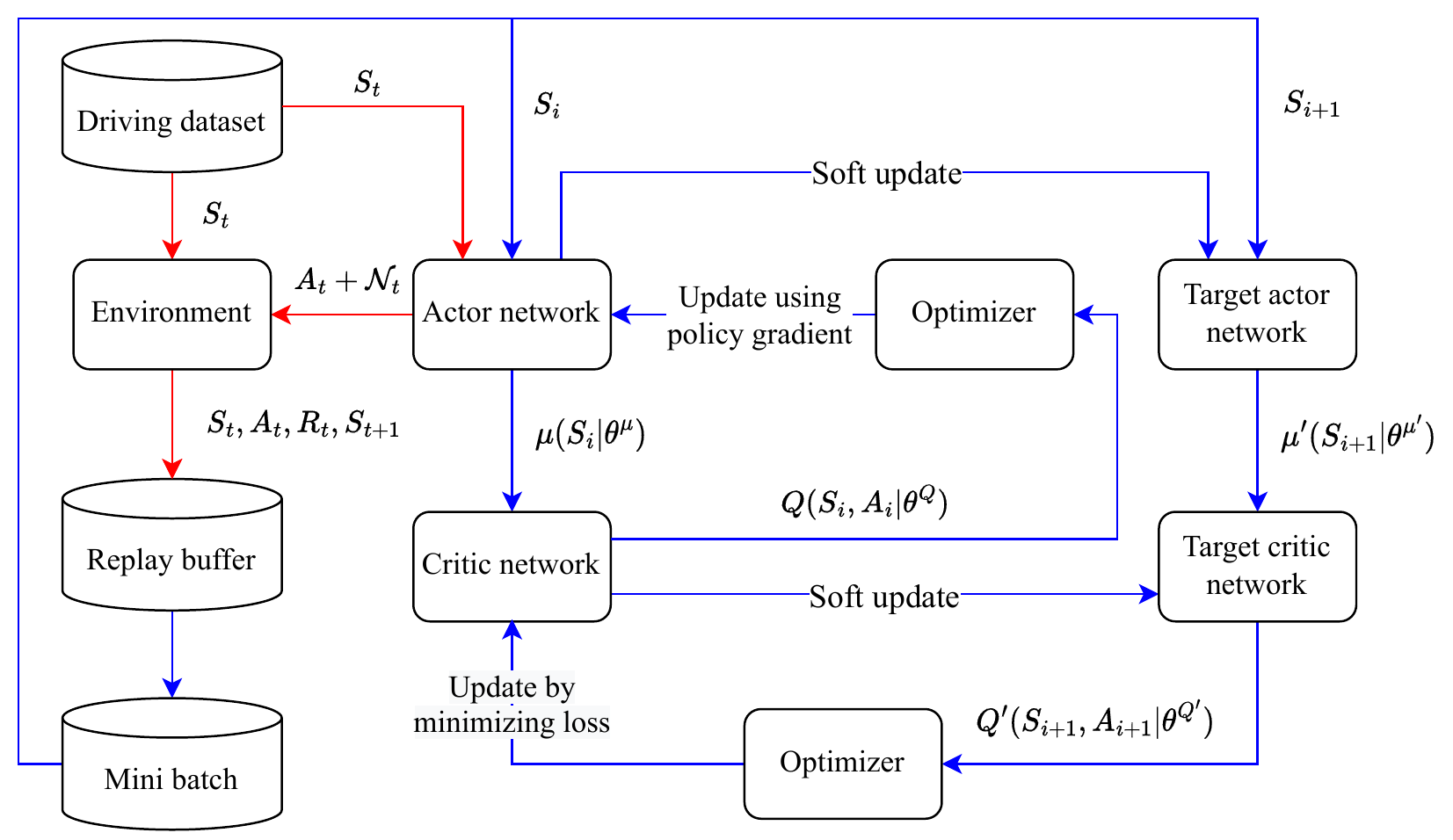}
\caption{The architecture of DDPG, including the experience generation (red arrows) and training (blue arrows) processes}
\label{ddpg_architecture}
\end{figure}

DDPG \citep{lillicrap2015continuous} is a model-free off-policy reinforcement learning algorithm, whose deep function approximators could capture complex non-linear relationships. It uses the actor-critic approach from DPG (deterministic policy gradient) \citep{silver2014deterministic} to deal with the policy in the continuous action space. Besides, it employs the experience replay and soft-updating target networks from DQN \citep{mnih2015human}, in order to reduce the correlation among samples and improve the learning efficiency and stability. The architecture of DDPG is illustrated in Fig. \ref{ddpg_architecture}, where the red arrows represent the experience generation process, and the blue arrows represent the training procedure.

The DDPG model consists of four neural networks, namely the actor ($\mu$), target actor ($\mu'$), critic ($Q$), and target critic ($Q'$) network. The actor networks ($\mu$ and $\mu'$) interact with the environment according to the given state and their weights. Then, the critic networks ($Q$ and $Q'$) evaluate the action and output $Q$-values, which would be used for the model update. The target networks ($\mu'$ and $Q'$) are the lagged version of the actual agent networks ($\mu$ and $Q$), which are used to improve the algorithm's stability.

In this study, the observed state $S_t=(d_{lon}(t), v_{lon}(t), \Delta v_{lon}(t), d_{lat}(t), v_{lat}(t), \Delta v_{lat}(t))$ at the time step $t$ is sampled from the naturalistic driving dataset, where $d_{lon}(t)$ and $d_{lat}(t)$ are the longitudinal and lateral distances to the leading vehicle, $v_{lon}(t)$ and $v_{lat}(t)$ are the CV's longitudinal and lateral speed, and $\Delta v_{lon}(t)$ and $\Delta v_{lat}(t)$ are the longitudinal and lateral relative speed, respectively. The action generated by the agent is $A_t=(\hat{a}_{lon}(t), \hat{a}_{lat}(t))$, which is composed of the longitudinal acceleration $\hat{a}_{lon}(t)$ and the lateral acceleration $\hat{a}_{lat}(t)$. Then, a noise process ($\mathcal{N}$) is employed to explore the state and action spaces beyond the dataset. The Ornstein-Uhlenbeck process ($\theta=0.15$ and $\sigma=0.2$) \citep{uhlenbeck1930theory} is used to generate the temporally correlated noise for a better exploration of physical environments with momentum. The sampled noise will be added to the action and used in the state update. In order to let the agent interact with the environment, a kinematic model is constructed for state updating, which is presented as

\begin{equation}\label{simulate}
\begin{aligned}
&\hat{v}_{lon}(t+1) = v_{lon}(t)+\hat{a}_{lon}(t)\Delta t \\
&\Delta \hat{v}_{lon}(t+1) = \Delta v_{lon}(t)-\hat{a}_{lon}(t)\Delta t\\
&\hat{d}_{lon}(t+1) = d_{lon}(t)+(v_{lon}(t)+\Delta v_{lon}(t))\Delta t-\frac{v_{lon}(t)+\hat{v}_{lon}(t+1)}{2} \Delta t \\
&\hat{v}_{lat}(t+1) = v_{lat}(t)+\hat{a}_{lat}(t)\Delta t\\
&\Delta \hat{v}_{lat}(t+1) = \Delta v_{lat}(t)-\hat{a}_{lat}(t)\Delta t\\
&\hat{d}_{lat}(t+1) = d_{lat}(t)+(v_{lat}(t)+\Delta v_{lat}(t))\Delta t-\frac{v_{lat}(t)+\hat{v}_{lat}(t+1)}{2} \Delta t
\end{aligned}
\end{equation}

where $\hat{v}_{lon}(t+1), \Delta \hat{v}_{lon}(t+1), \hat{d}_{lon}(t+1), \hat{v}_{lat}(t+1), \Delta \hat{v}_{lat}(t+1)$ and $\hat{d}_{lat}(t+1)$ are the simulated longitudinal speed, longitudinal relative speed, longitudinal distance, lateral speed, lateral relative speed and lateral distance at time step $t+1$, respectively. $\Delta t$ is the update time interval, which is 0.1s in this study. Three different reward functions, namely the distance reward ($R_d$), the speed reward ($R_v$), and the speed difference reward ($R_{\Delta v}$), are given by Eq. (\ref{r_s}), Eq. (\ref{r_v}), and Eq. (\ref{r_deltav}), respectively. A small $\epsilon$ value ($\epsilon = 1 \times 10^{-7}$) is added to avoid dividing by zero. The reward functions are designed to let the network imitate human drivers' behaviors and explore the underlying strategies by minimizing the difference in vehicle motions. At last, a set of experience ($S_t,A_t,R_t,S_{t+1}$) will be stored in the replay buffer.

\begin{equation}\label{r_s}
R_{d} = -(\frac{|\hat{d}_{lon}(t+1)-d_{lon}(t+1)|}{|d_{lon}(t+1)|+\epsilon})-(\frac{|\hat{d}_{lat}(t+1)-d_{lat}(t+1)|}{|d_{lat}(t+1)|+\epsilon})
\end{equation}

\begin{equation}\label{r_v}
R_{v} = -(\frac{|\hat{v}_{lon}(t+1)-v_{lon}(t+1)|}{|v_{lon}(t+1)|+\epsilon})-(\frac{|\hat{v}_{lat}(t+1)-v_{lat}(t+1)|}{|v_{lat}(t+1)|+\epsilon})
\end{equation}

\begin{equation}\label{r_deltav}
R_{\Delta v} = -(\frac{|\Delta \hat{v}_{lon}(t+1)-\Delta v_{lon}(t+1)|}{|\Delta v_{lon}(t+1)|+\epsilon})-(\frac{|\Delta \hat{v}_{lat}(t+1)-\Delta v_{lat}(t+1)|}{|\Delta v_{lat}(t+1)|+\epsilon})
\end{equation}

At the beginning of the model training process, a random mini-batch of $N$ transitions ($S_i,A_i,R_i,S_{i+1}$) are sampled from the replay buffer. The critic network evaluates the action and outputs the scalar $Q$-value $Q(S_i,A_i|\theta^Q)$ according to its weights $\theta^Q$. Similarly, the target actor network chooses the action $A_{i+1}=\mu'(S_{i+1}|\theta^{\mu'})$ based on the next-step state $S_{i+1}$, and the target critic network evaluates it with $Q'(S_{i+1},A_{i+1}|\theta^{Q'})$. The effect of the current action in the next step is taken into account in this step. Then, the critic network is updated by minimizing the loss $L=\frac{1}{N} \Sigma_i (y_i-Q(S_i,A_i|\theta^Q))^2$, where $y_i=R_i+\gamma Q'(S_{i+1},A_{i+1}|\theta^{Q'})$. The actor network is updated using the sampled policy gradient  $\nabla_{\theta\mu}J \approx \frac{1}{N} \sum_{i} \nabla_A Q(S,A|\theta^Q)|_{S=S_i, A=\mu(S_i)}\nabla_{\theta\mu}\mu(S|\theta^\mu)|_{S_i}$. In the end, the target actor and critic networks are soft updated with $\theta^{Q'} \gets \tau \theta^Q + (1-\tau)\theta^{Q'}$ and $\theta^{\mu'} \gets \tau \theta^\mu + (1-\tau)\theta^{\mu'}$, respectively. The detailed DDPG algorithm is presented in Algorithm \ref{ddpg_algotirhm}.

\begin{algorithm}[htpb]
    \caption{DDPG algorithm (based on \cite{lillicrap2015continuous})}\label{ddpg_algotirhm}
    Randomly initialize critic network $Q(S,A|\theta^Q)$ and actor $\mu(S|\theta^\mu)$ with weights $\theta^Q$ and $\theta^\mu$ \\
    Initialize target network $Q'$ and $\mu'$ with weights $\theta^{Q'} \gets \theta^Q$, $\theta^{\mu'} \gets \theta^\mu$ \\
    Initialize replay buffer $B$
    \begin{algorithmic}[0]
        \For{episode=1, M}
            \State Initialize a random process $\mathcal{N}$ for action exploration
            \State Receive initial observation state $S_1$
            \For{t=1, T}
                \State Select action $A_t=\mu(S_t|\theta^\mu)+\mathcal{N}_t$ according to the current policy and exploration noise
                \State Execute action $A_t$ and observe reward $R_t$ and observe new state $S _{t+1}$
                \State Store transition $(S_t, A_t, R_t, S_{t+1})$ in $B$
                \State Sample a random minibatch of $N$ transitions $(S_i, A_i, R_i, S_{i+1})$ from $B$
                \State Set $y_i=R_i+\gamma Q'(S_{i+1},\mu'(S_{i+1}|\theta^{\mu'})|\theta^{Q'})$
                \State Update critic by minimizing the loss: $L=\frac{1}{N} \Sigma_i (y_i-Q(S_i,A_i|\theta^Q))^2$
                \State Update the actor policy using the sampled policy gradient:\\
                \begin{center}
                    $\nabla_{\theta\mu}J \approx \frac{1}{N} \sum_{i} \nabla_A Q(S,A|\theta^Q)|_{S=S_i, A=\mu(S_i)}\nabla_{\theta\mu}\mu(S|\theta^\mu)|_{S_i}$
                \end{center}
                \State Update the target networks:\\
                \begin{center}
                    $\theta^{Q'} \gets \tau \theta^Q + (1-\tau)\theta^{Q'}$\\
                    $\theta^{\mu'} \gets \tau \theta^\mu + (1-\tau)\theta^{\mu'}$
                \end{center}
            \EndFor
        \EndFor
\end{algorithmic}
\end{algorithm}

\section{Data preparation}\label{data}




SPMD is the world's largest CV test program, which aims to demonstrate CV technologies in the real-world environment \citep{huang2017empirical}. 2,842 equipped vehicles participated in the program for over 2 years in Ann Arbor, Michigan \citep{bezzina2014safety}. The two-month sample data (October 2012 and April 2013) are now available to the public on the ITS Data Hub (\url{https://www.its.dot.gov/data/}). The SPMD environment includes eight datasets, namely Data Acquisition System 1 (DAS1), Data Acquisition System 2 (DAS2), Basic Safety Message (BSM), Roadside Equipment, Network, Weather, Schedule, and Road Work Activity \citep{hamilton2015safety}. 

The DAS1 dataset, collected by the University of Michigan Transportation Research Institute (UMTRI) in April 2013, was used in this study. A total of 7,960 trips recorded by 98 sedans equipped with the DAS1 and the MobilEye sensor \citep{harding2014vehicle} were investigated. Within the DAS1 dataset, the \textit{DataLane} file records the CVs' lateral positions relative to lane boundaries (\textit{LaneDistanceLeft} and \textit{LaneDistanceRight}) and the estimated lane marking measurement quality (\textit{LaneQualityLeft} and \textit{LaneQualityRight}). The \textit{DataFrontTargets} file provides relative positions (\textit{Range} and \textit{Transversal}) and relative speed (\textit{RangeRate}) of the leading vehicles. The \textit{DataWsu} file contains the geographical positions (\textit{LatitudeWsu} and \textit{LongitudeWsu}) and kinematics (\textit{GpsSpeedWsu} and \textit{AxWsu}) of CVs. The detailed description of the fields is reported in Table \ref{description}. All the data elements 
were collected at a frequency of 10 Hz. The \textit{DataLane} and \textit{DataFrontTargets} files were recorded by the MobilEye sensor, and the \textit{DataWsu} file was collected by the GPS unit and controller area network (CAN) bus via the wireless safety unit (WSU). Python programming language \citep{python} with Apache Spark bigdata analytic engine \citep{zaharia2016apache} was used for big data manipulation. 
\begin{table}[htpb]
\caption{Field description of CV data}
\label{description}
\centering
\small
\begin{tabular}{p{0.18\linewidth} p{0.18\linewidth} p{0.55\linewidth}}
\hline
Origin File               & Field Name                   & Description                                                                                                                                                 \\ \hline
Common                    & Device                       & A unique, numeric ID assigned to each DAS                                                                                                                   \\
                          & Trip                         & Count of ignition cycles—each ignition cycle commences when the ignition is in the on position and ends when it is in the off position                      \\
                          & Time                         & Time in centiseconds since DAS started, which (generally) starts when the ignition is in the on position ($centisecond$)                                    \\
\textit{DataLane}         & LaneDistanceLeft ($d_{l}$)   & Distance between the center line of the vehicle and the left boundary of the travel lane ($m$)                                                              \\
                          & LaneDistanceRight ($d_{r}$)  & Distance between the center line of the vehicle and the right boundary of the travel lane ($m$)                                                             \\
                          & LaneQualityLeft              & Quality of the estimated boundary measure of the travel lane’s left boundary (ranging from 0 "very bad" to 3 "very good")                                   \\
                          & LaneQualityRight             & Quality of the estimated boundary measure of the travel lane’s left boundary (ranging from 0 "very bad" to 3 "very good")                                   \\
\textit{DataFrontTargets} & ObstacleId                   & ID of new obstacle, as assigned by the Mobileye sensor, and its value will be the last used free ID                                                         \\
                          & TargetType                   & Classification of an identified obstacle/target (0: car; 1: truck; 2: motorcycle; 3: pedestrian; 4: bicycle)                                                \\
                          & Range ($d_{lon}$)           & Longitudinal position of an object, typically the closest object, relative to a reference point on the host vehicle, according to the Mobileye sensor ($m$) \\
                          & RangeRate ($\Delta v_{lon}$) & Longitudinal velocity of an object, typically the closest object, relative to the host vehicle, according to the Mobileye sensor ($m/s$)                    \\
                          & Transversal ($d_{lat}$)     & The lateral position of the obstacle, as determined by the Mobileye sensor ($m$)                                                                            \\
\textit{DataWsu}          & GpsValidWsu                  & Communicates whether a GPS data point is valid (1) or not (0)                                                                                               \\
                          & LatitudeWsu                  & Latitude from WSU receiver ($deg$)                                                                                                                          \\
                          & LongitudeWsu                 & Longitude from WSU receiver ($deg$)                                                                                                                         \\
                          & GpsSpeedWsu ($v_{lon}$)     & Speed from WSU GPS receiver ($m/s$)                                                                                                                         \\
                          & ValidCanWsu                  & Vehicle CAN Bus message to WSU is valid (1) or not(0)                                                                                                       \\
                          & AxWsu                        & Longitudinal acceleration from vehicle CAN Bus vis WSU ($m/s^2$)                                                                                             \\ \hline
\end{tabular}
\end{table}

To begin with, the invalid records were filtered out from the dataset. The criterion "\textit{LaneQualityLeft > 0}  and \textit{LaneQualityRight > 0}" was used to remove records with poor distance measurement quality. The data points with valid GPS and Can Bus messages were extracted with the criterion "\textit{GpsValidWsu = 1} and \textit{ValidCanWsu = 1}". As the MobilEye sensor can track and identify different types of front objects, the criterion "\textit{TargetType = 0}" was employed to ensure the leading vehicles are cars. After the cleaning process, all three datasets were merged based on the common fields\textit{Device}, \textit{Trip}, and \textit{Time}.

Then, some records were filtered out based on their values. According to the technical report \citep{kbb2013}, the fastest sedan can travel at a speed of 200mph (90m/s), and the maximum acceleration is around $\rm7m/s^2$ in 2013. The record with \textit{GpsSpeedWsu} value higher than 90m/s, or \textit{AxWsu} value greater than $\rm7m/s^2$ was removed as outliers. The MobilEye sensor could cover three or more lanes and track multiple targets, including the parking and reversing vehicles on the roadside, as well as the vehicles in the opposite direction. So the speed of the leading vehicle, calculated by \textit{GpsSpeedWsu + Rangerate}, needs to be examined. Considering that the reversing vehicle may have potential conflict with the ego vehicle, the criterion "\textit{GpsSpeedWsu + Rangerate > -1m/s}" was employed to filter out the records describing vehicles in the opposite direction. Besides, the free-flow scenarios were eliminated with the rule "\textit{Range < 100m}". The criterion "\textit{-7m < Transversal < 7m}" was applied to ensure that the leading vehicle is in the same or adjacent lane of the subject vehicle.

\begin{figure}[!htpb]
\centering\includegraphics[width=0.6\linewidth]{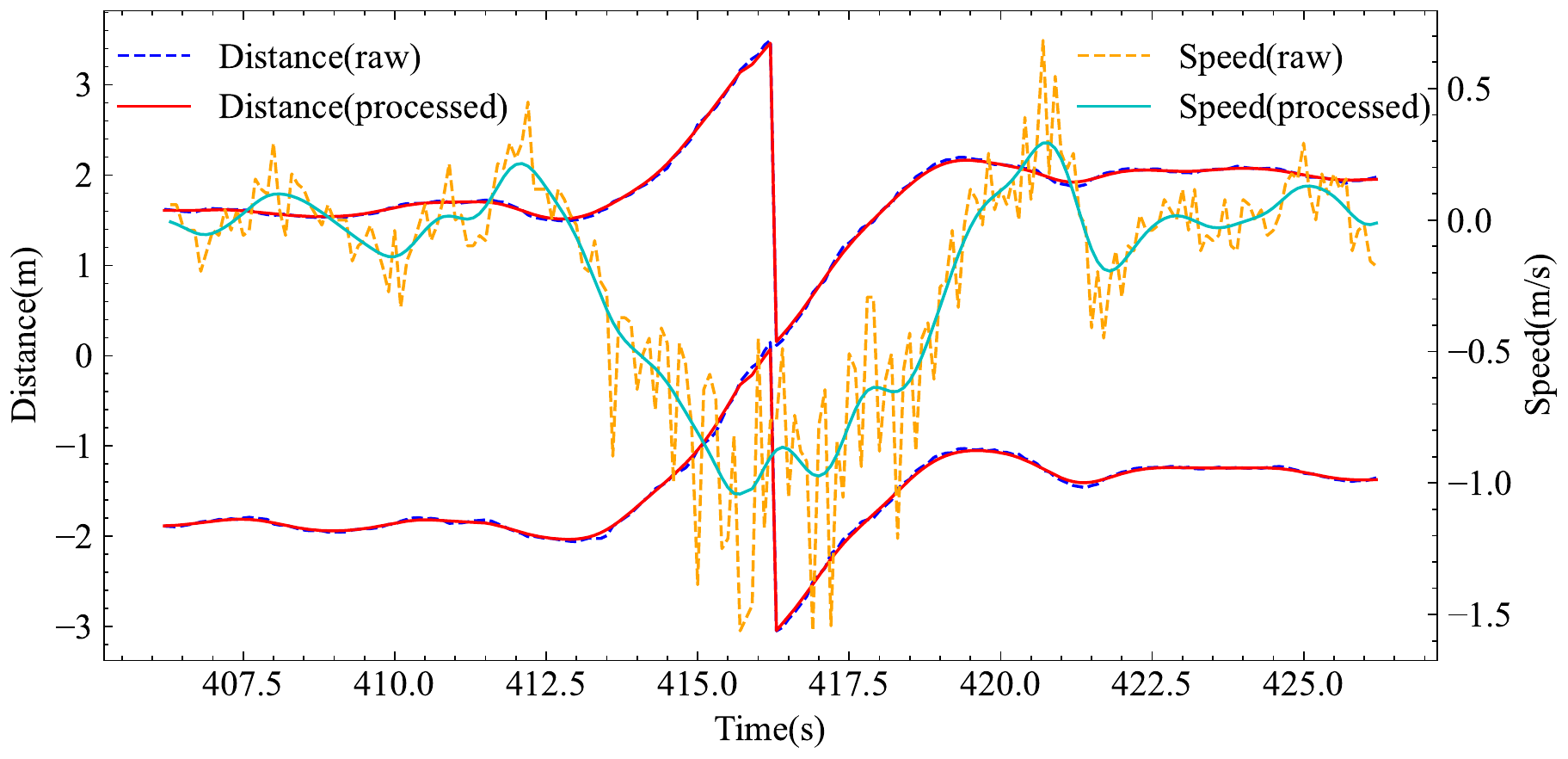}
\caption{The raw and processed lateral distance and speed data}\label{gaussian_filter}
\end{figure}

Next, the data would be filtered to reduce the influence of noise. The measurement quality of lane distance data ($LaneDistanceLeft$ and $LaneDistanceRight$) is not satisfactory, and the estimated quality ($LaneQualityLeft$ and $LaneQualityRight$) is usually less than 3. Because the lane markings might be uncontinuous and unclear, and they could be shaded by other vehicles. A Gaussian filter was adopted to process the lane distance data for noise reduction. The raw and processed data are shown in Fig \ref{gaussian_filter}. The computation of lateral speed is based on the lane distance data, which is given by Eq. \ref{vlat}. It is obvious that the Gaussian filter can effectively reduce the fluctuation of lateral speed without changing the lateral distance significantly.

\begin{figure}[!htpb]
\centering\includegraphics[width=1\linewidth]{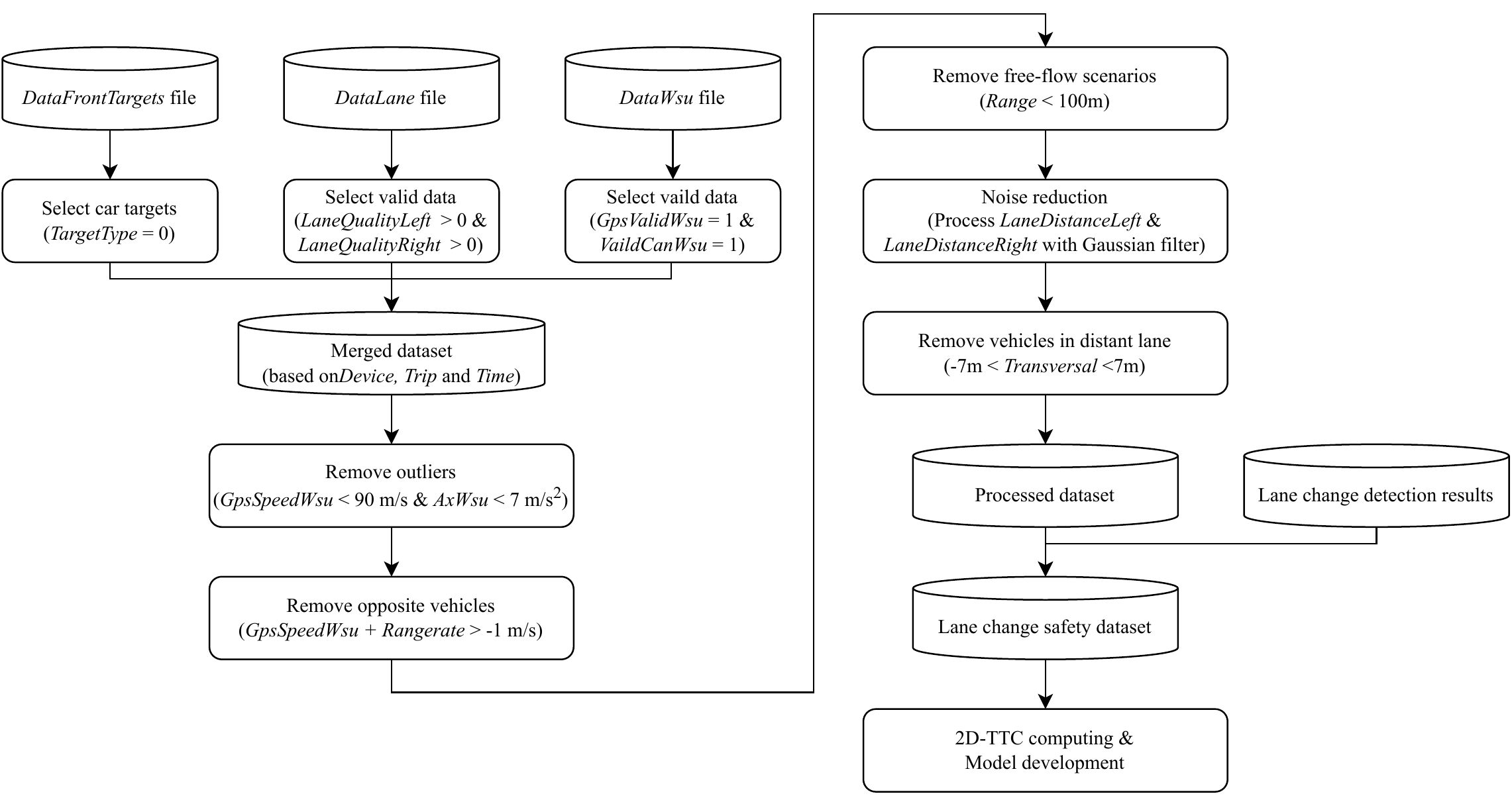}
\caption{Data preparation procedure}\label{data_preparation}
\end{figure}

At last, the driving scenarios related to lane change were extracted. From the impact of the lane change on the traffic flow perspective, it lasts for about 25s on average \citep{ma2008comparisons,wang2008effect}. \cite{zheng2013effects} pointed out that the average duration for the anticipation and relaxation processes of lane change is 8-14s and 10-15s, respectively. To cover the complete process of lane change maneuver, the lane change events were selected 15s before and 15s after the time step of the vehicle crossing the lane boundary. Based on the findings in \cite{guo2022lane}, 2,199 left and 1,663 right lane change samples were detected from the SPMD dataset, which will be used for 2D-TTC computation and evasive behavior model development. The detailed data preparation procedure is illustrated in Fig. \ref{data_preparation}.

In the processed lane change safety dataset, the longitudinal speed of the ego vehicle $v_{lon} = GpsSpeedWsu$ and the relative lateral distance $s_{0lat} = Transversal$ can be exported. Other variables in the 2D-TTC equations are calculated as

\begin{equation}
    s_{0lon} = d_{lon}+l
\end{equation}

\begin{equation}\label{vlat}
    v_{lat} = \frac{(d_{l}(t)-d_{l}(t-1))+(d_{r}(t)-d_{r}(t-1))}
    {2\times\Delta t} 
\end{equation}

\begin{equation}
    v_{0lon} = v_{lon} + \Delta v_{lon} 
\end{equation}

\begin{equation}
    v_{0lat} = v_{lat} + \frac{s_{0lat}(t) - s_{0lat}(t-1)}{\Delta t}
\end{equation}

where $\Delta t$ is the time interval between two continuous records, which is 0.1s in the dataset. The length and width of leading cars are assumed to be 4.8m and 1.6m, based on the average car size.

\section{Results and discussion}\label{results}

\subsection{Validation of 2-D TTC}\label{ttc_val}

It is necessary to confirm the validity of 2D-TTC before using it for potential risk detection. The method computing the correlation coefficients between the risks captured by 2D-TTC and the archived crashes was employed for SSM validation \citep{xie2019use}. A total of 75 highway segments with complete traffic data were selected from the Ann Arbor road network for analysis. The traffic volume data of these highways were obtained from the Michigan Department of Transportation (MDOT) Open Data (\url{http://gis-mdot.opendata.arcgis.com/}), and the crash data were gained from the Michigan's Open Data Portal (\url{https://data.michigan.gov/}). ArcGIS \citep{arcgis} was used for spatial data processing in this section.

The GPS points collected by CVs and the recorded crashes were assigned to the road segments if the distance to the nearest road is less than 10m. After processing, 1,472 lane change scenarios, 1,025 rear-end crashes, and 336 sideswipe crashes were allocated to the selected highways. Multiple threshold values for the 2D-TTC were tested with an increment of 0.1s. Each GPS point would be given a binary risk value (0: 2D-TTC value is higher than the threshold, and 1: otherwise) based on the current threshold. Then, the risks and crashes were aggregated on each road segment $i$, and the aggregated variables are denoted as $RiskCount_i$ and $CrashCount_i$, respectively. To account for the confounding effect of exposure indicators, the rates of risk and crash on the road segment would be used for correlation analysis. The $RiskRate_i$ and $CrashRate_i$ are calculated by Eq. \ref{riskrate} and \ref{crashrate}, respectively.

\begin{equation}\label{riskrate}
    RiskRate_i=\frac{RiskCount_i}{GPSCount_i}
\end{equation}

\begin{equation}\label{crashrate}
    CrashRate_i=\frac{CrashCount_i}{AADT_i}
\end{equation}

where $i$ is the road segment index, $GPSCount_i$ is the amount of GPS points on road segment $i$, and $AADT_i$ is the annual average daily traffic volume of the road segment $i$.

\begin{figure}[!htpb]
    \centering
    \begin{subfigure}{0.3\linewidth}
        \centering
        \includegraphics[width=1\textwidth]{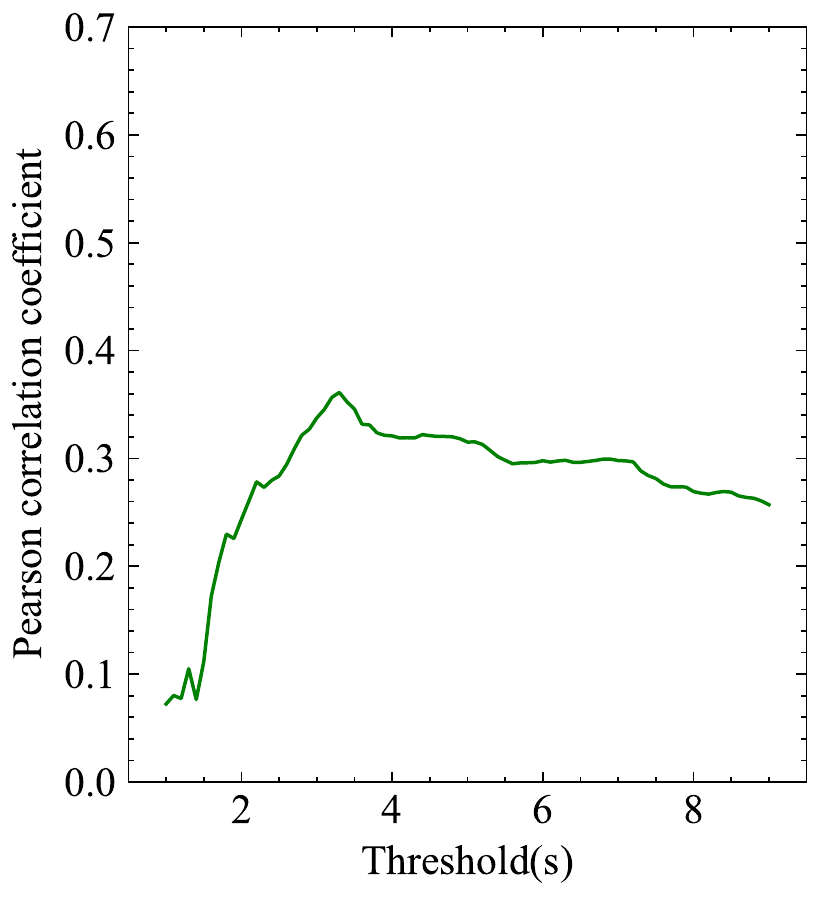}
        \caption{Rear-end risk vs rear-end crash}
    \end{subfigure}
    \begin{subfigure}{0.3\linewidth}
        \centering
        \includegraphics[width=1\textwidth]{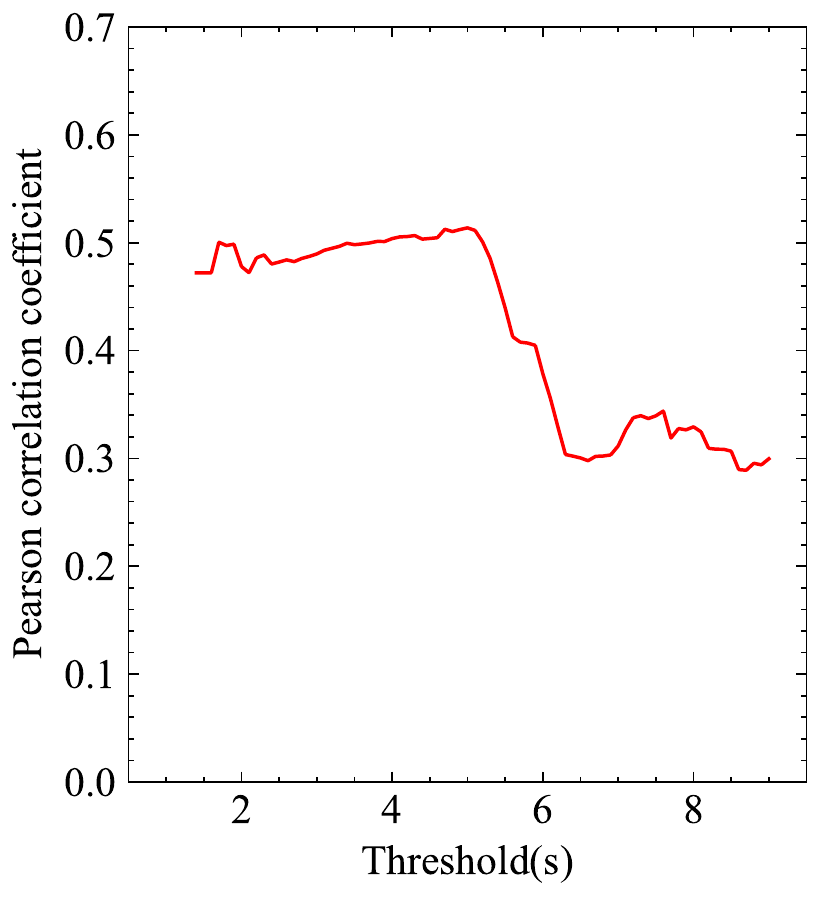}
        \caption{Sideswipe risk vs sideswipe crash}
    \end{subfigure}
    \begin{subfigure}{0.3\linewidth}
        \centering
        \includegraphics[width=1\textwidth]{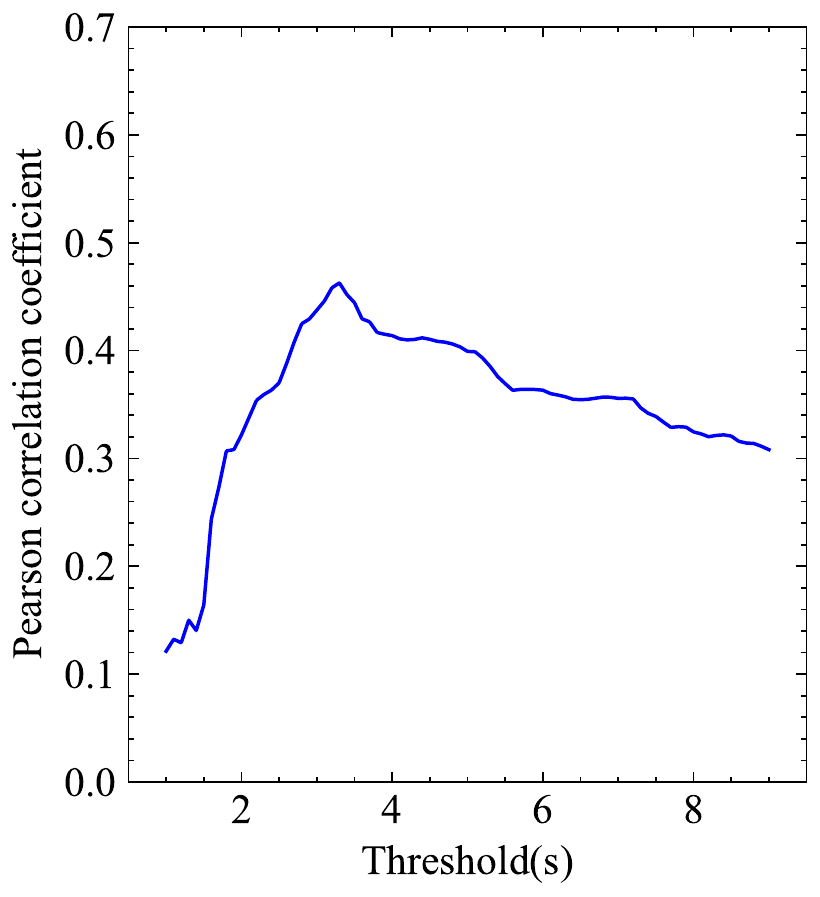}
        \caption{All risk vs all crash}
    \end{subfigure}
\caption{Correlation coefficient between risk rate and crash rate with different combinations}\label{ttc_pearson}
\end{figure}

For each threshold value, the Pearson correlation coefficient \citep{pearson1895vii} between $RiskRate_i$ and $CrashRate_i$ was computed. Three combinations of risk and crash types were tested, namely rear-end risks vs rear-end crashes, sideswipe risks vs sideswipe crashes, and rear-end and sideswipe (all) risks vs rear-end and sideswipe (all) crashes. The results are shown in Fig. \ref{ttc_pearson}, and the details of the correlation test with the highest correlation coefficient are presented in Table \ref{correlation_test}.

\begin{table}[!htpb]
\centering
\caption{Correlation test results}
\label{correlation_test}
\begin{tabular}{lllll}
\hline
Risk type      & Crash type     & Optimal threshold(s) & Pearson correlation coefficient & P-value \\ \hline
Rear           & Rear           & 3.3              & 0.361                           & 0.0015  \\
Sideswipe      & Sideswipe      & 5.0              & 0.514                           & 0.0000  \\
Rear+sideswipe & Rear+sideswipe & 3.3              & 0.463                           & 0.0000  \\ \hline
\end{tabular}
\end{table}

It is clear that the risks captured during the lane change duration are closely correlated with sideswipe crashes, which are more frequent than rear-end crashes in lane-change-related crashes \citep{sen2003analysis}. The highest correlation coefficient 0.514 is obtained between the captured sideswipe risks and the archived sideswipe crashes since the sideswipe crash is closely correlated with the lane change maneuvers. As shown in Table \ref{correlation_test}, all three p-values are less than 0.05, which indicates the statistically significant correlation between the 2D-TTC and the crashes. The thresholds are also within a reasonable range of common TTC thresholds \citep{pinnow2021review}. Therefore, the 2D-TTC is a reliable SSM to capture the potential risks in lane change scenarios.

\subsection{DDPG model training and results}

In this section, the DDPG model was trained with the driving data of safety-critical lane changes. An appropriate pre-specified threshold for 2D-TTC is required to identify and extract the conflicts during lane change maneuvers. In the ADAS applications, the threshold of TTC is usually 5s \citep{van1993time, scott2008comparison, mohebbi2009driver, biondi2017advanced}. The warning is presented as long as the TTC is less than 5s, and the driver is reminded to take evasive actions. Additionally, evasive actions are often found to be executed within a second after the TTC decreases below 5s in the historical (near) crash data \citep{markkula2012review}. Considering the purpose of this study is to capture and model drivers' evasive behavior in safety-critical situations, the threshold for 2D-TTC was set to 5s. If there are more than 10 continuous records with 2D-TTC values less than the threshold, it would be regarded as a conflict. After identification and extraction, there were 903 conflicts (lasting for 17,699 time steps) in 743 lane change events. 80\% of the data will be used for model training, and the rest 20\% are used for model testing. 

\begin{figure}[!htpb]
    \centering
    \includegraphics[width=0.9\linewidth]{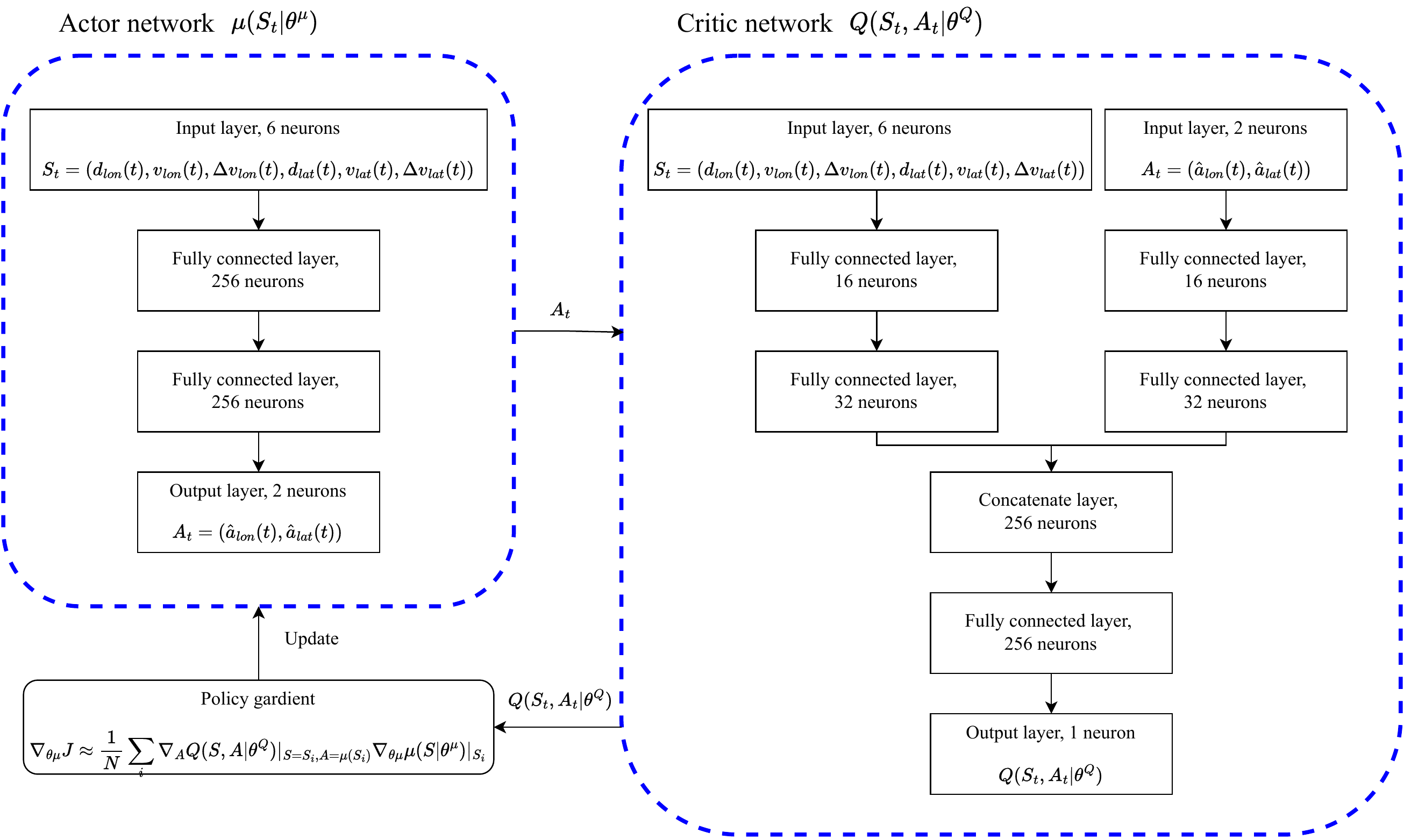}
\caption{Architectures of the actor and critic networks}\label{ac_architecture}
\end{figure}

Three DDPG models, namely the distance (DDPGd), speed (DDPGv), and speed difference (DDPG$\Delta$v) models, were trained using the distance ($R_d$), speed ($R_v$), and speed difference ($R_{\Delta v}$) rewards, respectively. The input of these models is the observed state at each time step $S_t=(d_{lon}(t), v_{lon}(t), \Delta v_{lon}(t), d_{lat}(t), v_{lat}(t), \Delta v_{lat}(t))$, and the outputs of the actor network are the vehicle's longitudinal and lateral accelerations ($\hat{a}_{lon}$ and $\hat{a}_{lat}$). The DDPG models are expected to learn the underlying policy of driver's evasive behavior through the interaction with the environment. The architectures of the actor and critic networks are illustrated in Fig. \ref{ac_architecture}. The target actor and critic networks have the same architecture as the actual agent actor and critic networks. The hyperparameters adopted for the DDPG model training are presented in Table \ref{hyperparameters}. Additionally, a neural network (NN) model with two 256-neuron hidden layers was developed as the benchmark for comparison. Its input is the same observed state $S_t$ as the DDPG models, and its outputs are also the vehicle's longitudinal and lateral accelerations ($\hat{a}_{lon}$ and $\hat{a}_{lat}$). The NN model is supposed to imitate the driver's evasive behaviors in the training set. 

\begin{table}[htpb!]
\centering
\caption{Hyperparameters in DDPG model}
\label{hyperparameters}
\begin{tabular}{lll}
\hline
Hyperparameter                   & Value  & Description                                                \\ \hline
Actor learning rate              & 0.0005 & Learning rate used by the Adam optimizer of actor network  \\
Critic learning rate             & 0.001  & Learning rate used by the Adam optimizer of critic network \\
Discount factor ($\gamma$)       & 0.9    & Discount factor in DDPG    \\
Soft target update rate ($\tau$) & 0.01   & Soft update rate for target networks                       \\
Replay memory size               & 10,000 & Number of training samples in replay memory                \\
Batch size                       & 256    & Number of training samples used for gradient update        \\ \hline
\end{tabular}
\end{table}

\begin{figure}[!htpb]
    \centering
    \begin{subfigure}{0.3\linewidth}
        \centering
        \includegraphics[width=1\textwidth]{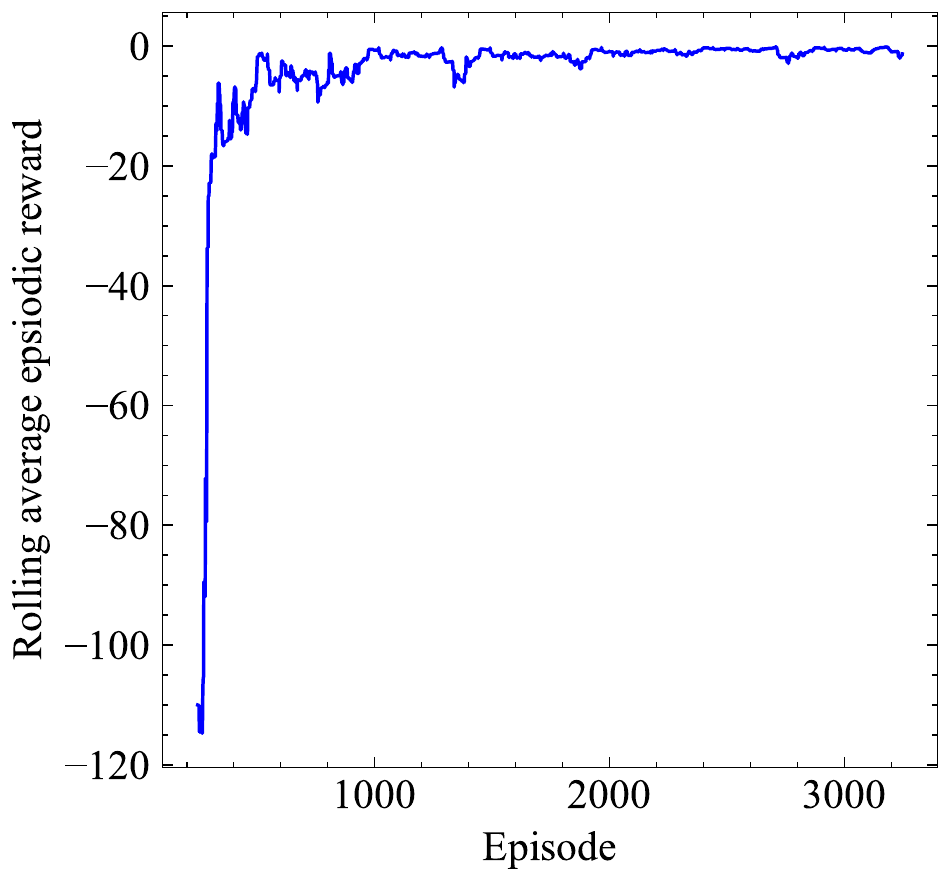}
        \caption{DDPGd model}\label{DDPGd_reward}
    \end{subfigure}
    \begin{subfigure}{0.3\linewidth}
        \centering
        \includegraphics[width=1\textwidth]{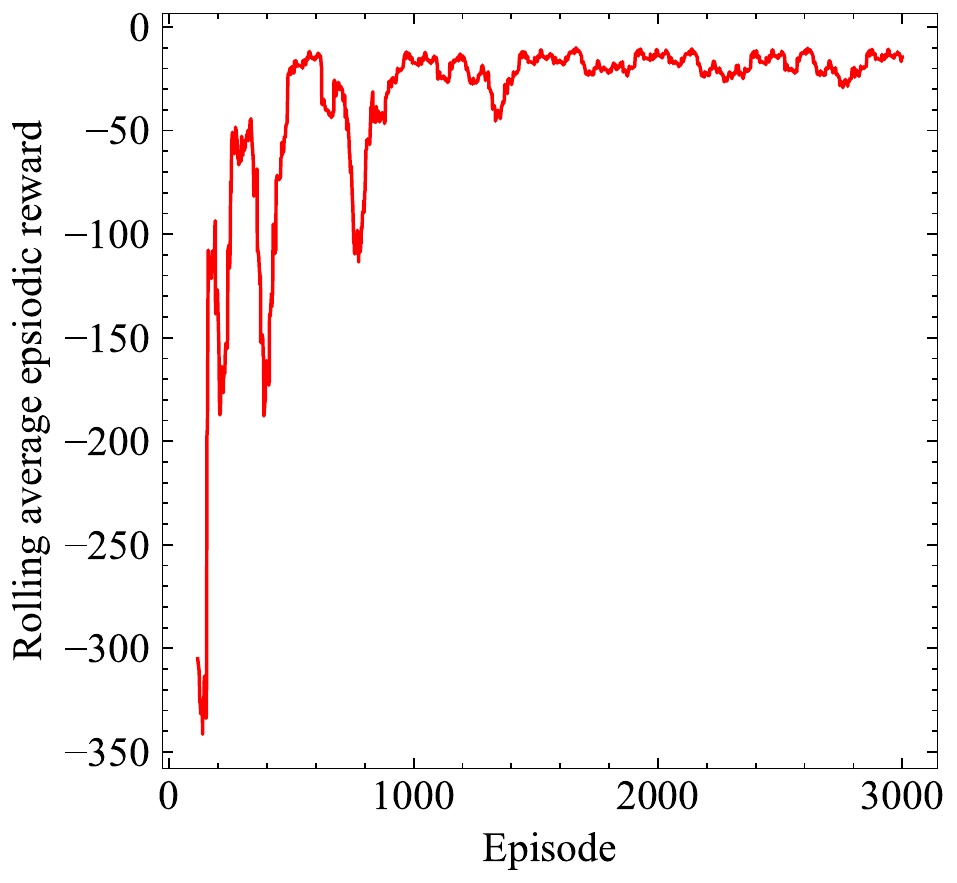}
        \caption{DDPGv model}\label{DDPGv_reward}
    \end{subfigure}
    \begin{subfigure}{0.3\linewidth}
        \centering
        \includegraphics[width=1\textwidth]{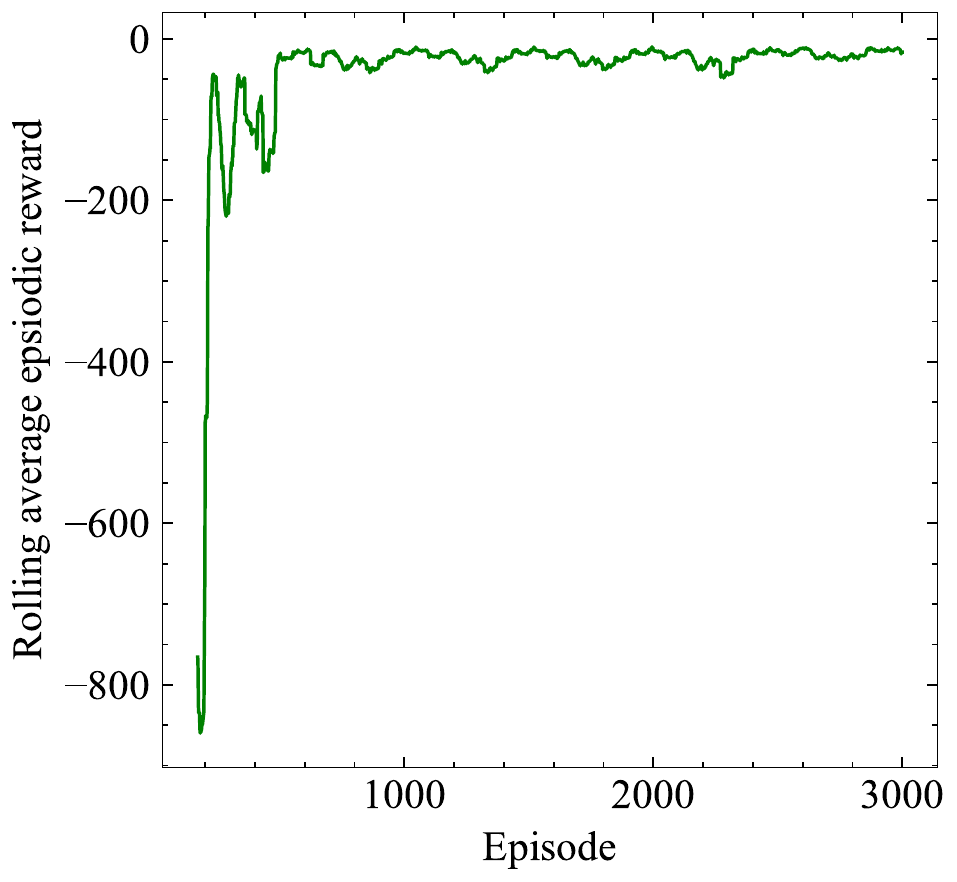}
        \caption{DDPG$\Delta$v model}\label{DDPGdv_reward}
    \end{subfigure}
    \caption{Rolling average episodic reward of DDPG models on training data}
    \label{reward}
\end{figure}

At the training stage, a safety-critical lane change event was sampled from the training set without replacement and passed to the DDPG model. The model would process the input event sequentially and store the set of experience ($S_t, A_t, R_t, S_{t+1}$) at each time step $t$ in the replay buffer. Then, another event was selected and processed by the DDPG model. If all the events have been used, the training set would be initialized. The training process is repeated for 3,000 episodes, where an episode means loading a safety-critical lane change event. The average episodic reward of the training set with a rolling window of 50 episodes is shown in Fig. \ref{reward}. All the models start to converge after around 1,500 episodes. The fluctuation at the converged stage is caused by the exploration noises.

The root mean square error (RMSE) and the Jensen–Shannon divergence (JSD) are employed to evaluate the performances of models in both the longitudinal and lateral motions. The RMSE computes the average error between the simulated and observed values as given by Eq \ref{rmse}.

\begin{equation}\label{rmse}
    RMSE = \sqrt{\frac{1}{n}\sum_{i=1}^{n}(\hat{x}_i-x_i)^2}
\end{equation}

where $\hat{x}_i$ and $x_i$ are the estimated and observed values, respectively. The RMSE of distance ($d_{lon}$ and $d_{lat}$), speed ($v_{lon}$ and $v_{lat}$) and acceleration ($a_{lon}$ and $a_{lat}$) are reported in Table \ref{model_performance}. The JSD is used to measure the similarity between the distributions and is also reported in Table \ref{model_performance}, which is calculated as

\begin{equation}\label{js}
    JSD(P||Q) = \frac{1}{2}\sum_{x\in \mathcal{X}}P(x)log(\frac{P(x)}{\frac{1}{2}(P(X)+Q(X)})+\frac{1}{2}\sum_{x\in \mathcal{X}}Q(x)log(\frac{Q(x)}{\frac{1}{2}(P(X)+Q(X)})
\end{equation}

where $P$ and $Q$ are the estimated and actual discrete probability distributions, respectively.

\begin{table}[!htpb]
\centering
\caption{Root mean square error (RMSE) and Jensen–Shannon divergence (JSD) between model results and real data}
\label{model_performance}
\begin{threeparttable}
\begin{tabular}{lllllllll}
\hline
\multirow{2}{*}{Variable} & \multicolumn{2}{l}{DDPGd} & \multicolumn{2}{l}{DDPGv}         & \multicolumn{2}{l}{DDPG$\Delta$v} & \multicolumn{2}{l}{NN} \\ \cline{2-9} 
                          & RMSE        & JSD         & RMSE            & JSD             & RMSE                 & JSD        & RMSE       & JSD       \\ \hline
$d_{lon}$                 & 0.0726      & 0.0006      & 0.0725* & 0.0006* & 0.0726               & 0.0006     & 0.0728     & 0.0006    \\
$v_{lon}$                 & 0.0478      & 0.0008      & 0.0425* & 0.0006* & 0.0481               & 0.0009     & 0.0539     & 0.0011    \\
$a_{lon}$                 & 0.4782      & 0.0182      & 0.4254* & 0.0134* & 0.4814               & 0.0216     & 0.5392     & 0.0258    \\
$d_{lat}$                 & 0.0013      & 0.0013      & 0.0012          & 0.0012* & 0.0010*     & 0.0014     & 0.0013     & 0.0033    \\
$v_{lat}$                 & 0.0229      & 0.0042      & 0.0223* & 0.0037* & 0.0232               & 0.0038     & 0.0235     & 0.0047    \\
$a_{lat}$                 & 0.2292      & 0.0138      & 0.2227* & 0.0113* & 0.2322               & 0.0190     & 0.2354     & 0.0198    \\ \hline
\end{tabular}
\begin{tablenotes}
\footnotesize
\item[*] The best values among all the models.
\end{tablenotes} 
\end{threeparttable}
\end{table}

In general, the proposed DDPG models could capture evasive behaviors accurately. Compared with the baseline NN model, they can achieve better performances in both RMSE and JSD. Because the DRL algorithm can learn the inherent mechanism of the evasive behaviors rather than only imitating them. The NN model becomes less accurate while handling the unseen scenarios in the test set.

\begin{figure}[!htpb]
    \centering
    \begin{subfigure}{0.32\linewidth}
        \centering
        \includegraphics[height=5cm]{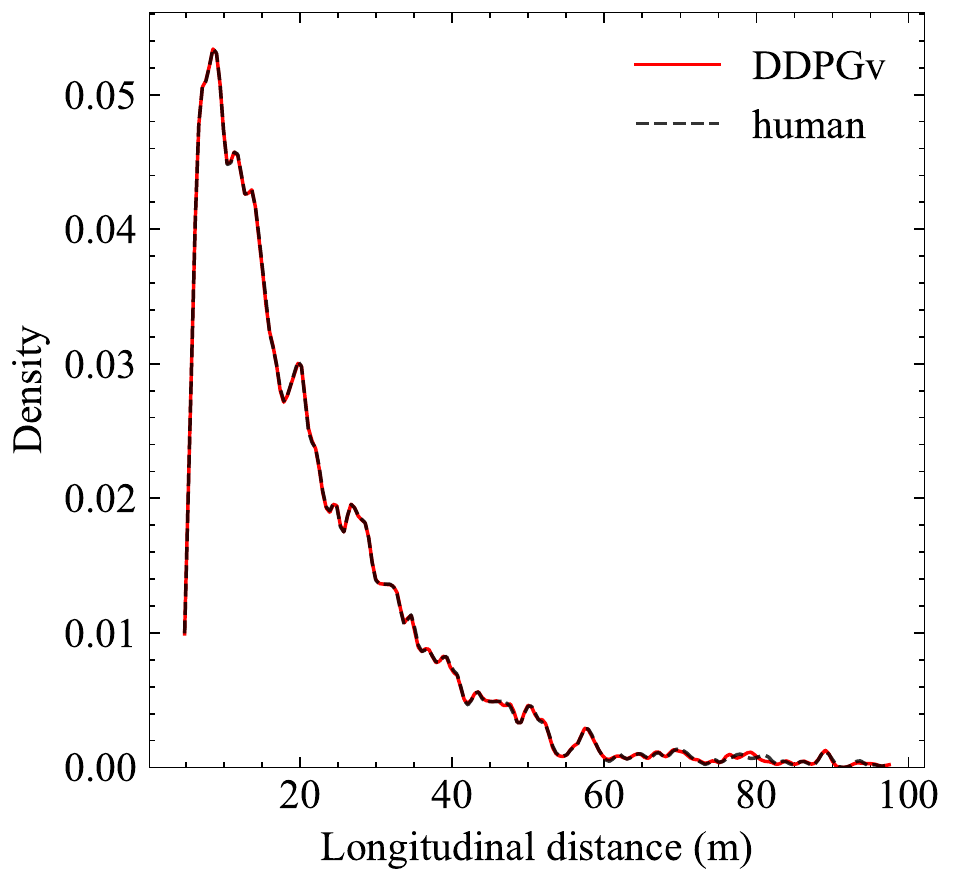}
        \caption{Distribution of longitudinal distance}\label{lon_d}
    \end{subfigure}
    \begin{subfigure}{0.32\linewidth}
        \centering
        \includegraphics[height=5cm]{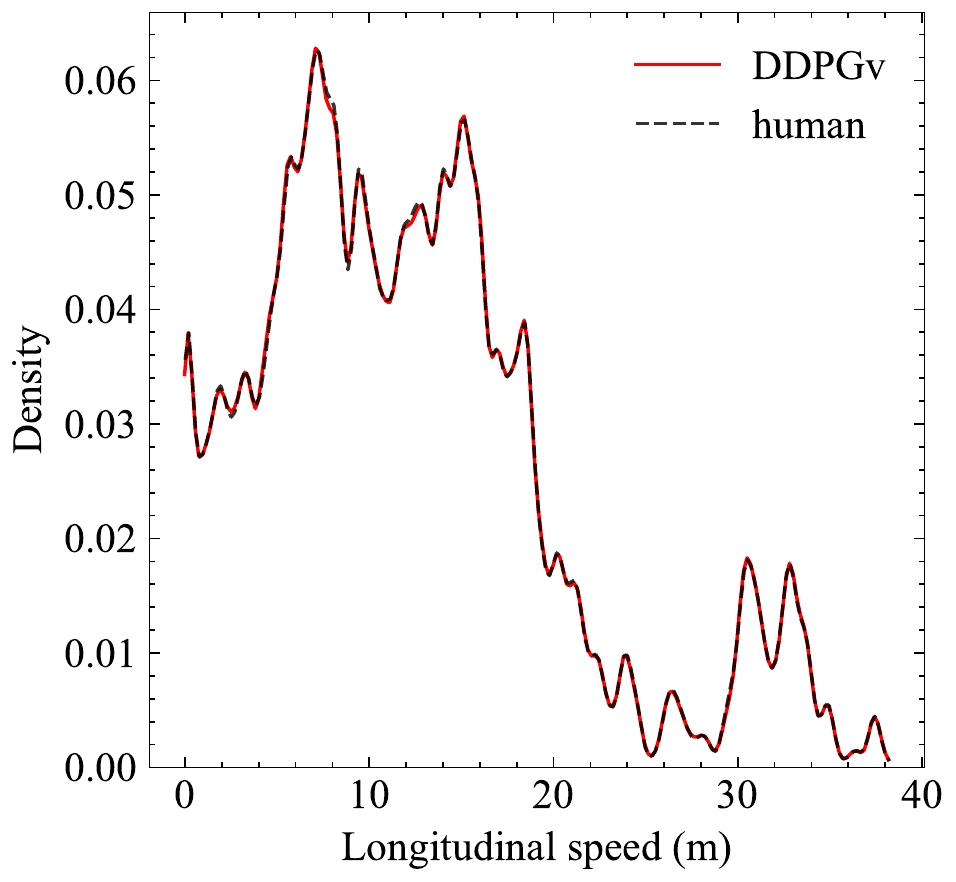}
        \caption{Distribution of longitudinal speed}\label{lon_v}
    \end{subfigure}
    \begin{subfigure}{0.32\linewidth}
        \centering
        \includegraphics[height=5cm]{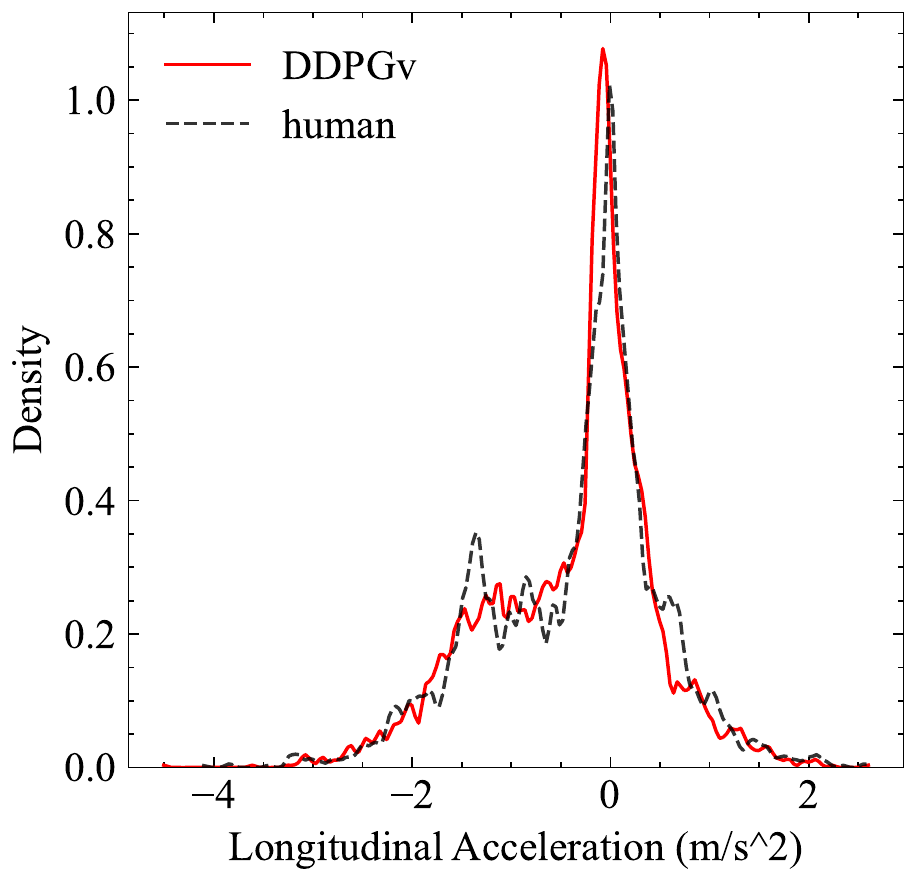}
        \caption{Distribution of longitudinal acceleration}\label{lon_a}
    \end{subfigure}
    
    \begin{subfigure}{0.32\linewidth}
        \centering
        \includegraphics[height=5cm]{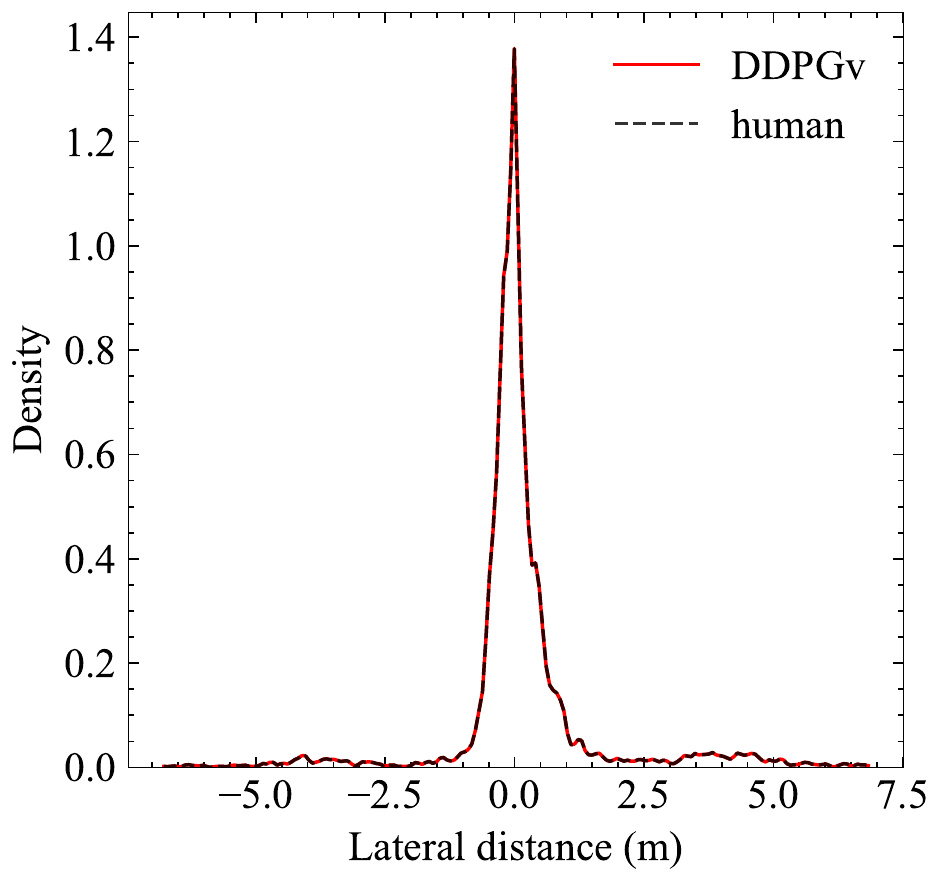}
        \caption{Distribution of lateral distance}\label{lat_d}
    \end{subfigure}
    \begin{subfigure}{0.32\linewidth}
        \centering
        \includegraphics[height=5cm]{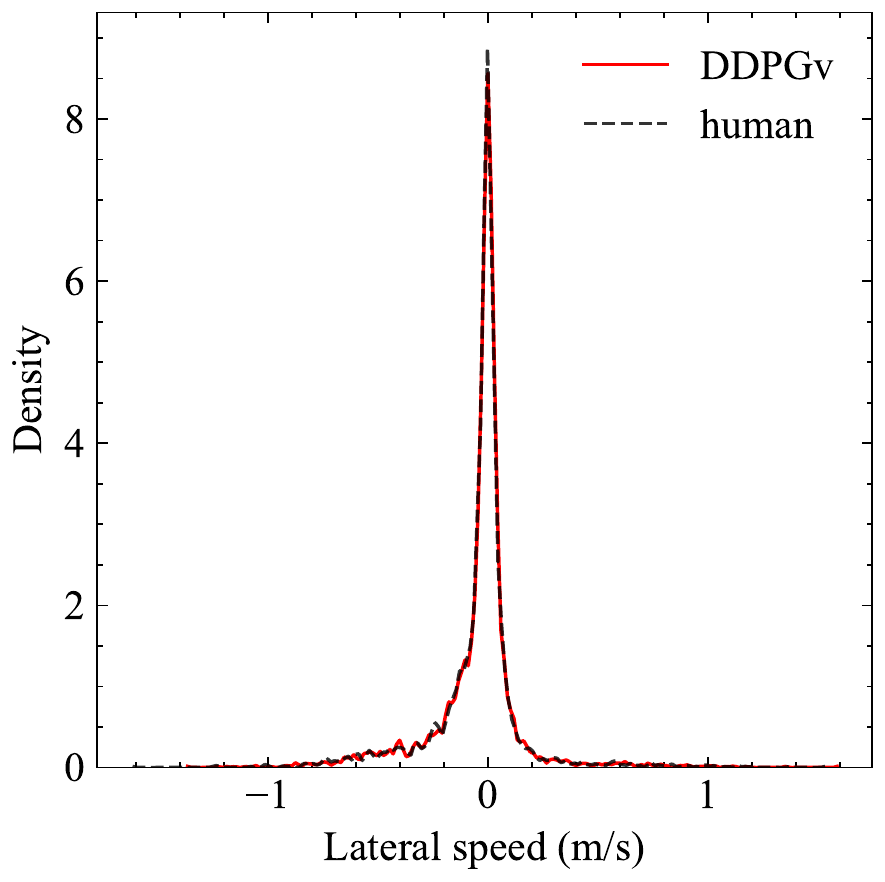}
        \caption{Distribution of lateral speed}\label{lat_v}
    \end{subfigure}
    \begin{subfigure}{0.32\linewidth}
        \centering
        \includegraphics[height=5cm]{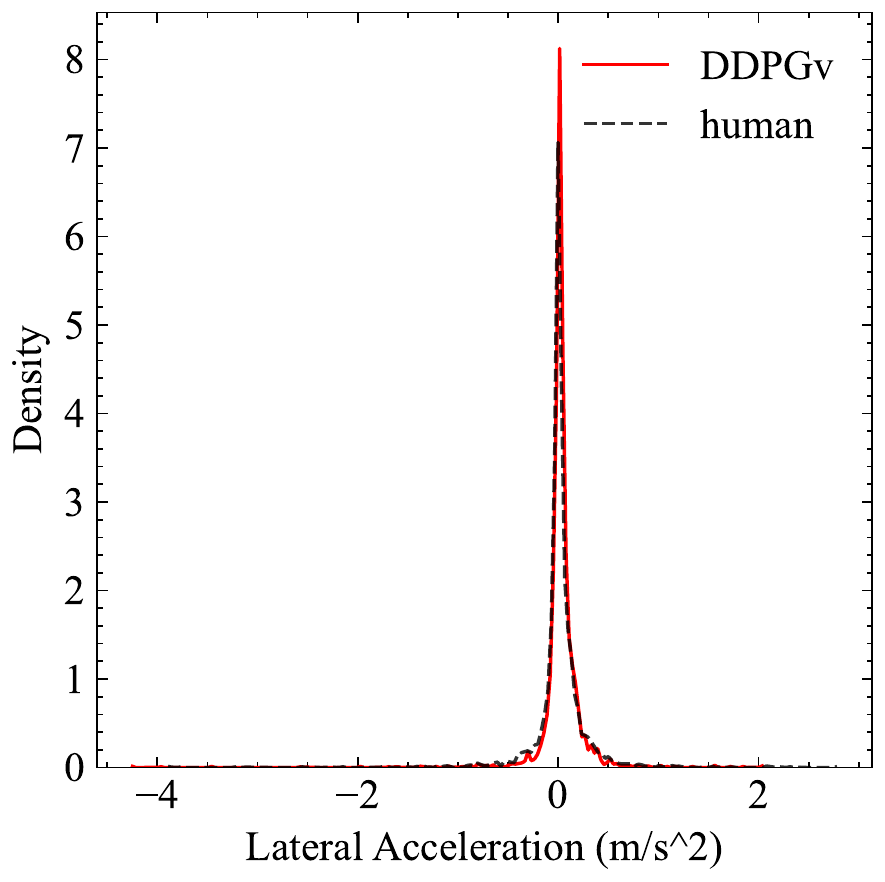}
        \caption{Distribution of lateral acceleration}\label{lat_a}
    \end{subfigure}
    \caption{Distributions of real and estimated values}
    \label{distribution}
\end{figure}

Among the DDPG models, the DDPGv model has the lowest RMSE and JSD values on most of the evaluation variables. It outperforms the other two models, especially in terms of longitudinal acceleration. The distributions of the values generated by the DDPGv model are compared with the real data distributions, as shown in Fig. \ref{distribution}. The distributions of the longitudinal and lateral distances, speeds, and accelerations generated by the model are highly consistent with the observed data. 

A possible reason for this performance difference is related to the state updating function, which is given by Eq. \ref{simulate}. The speed of the leading vehicle is assumed to be constant between the observed time $t$ and the simulated time $t+1$. However, the leading vehicle might change its speed in a real situation. The error between the simulated and real values is propagated to the models through the distance and speed difference reward functions (Eq. \ref{r_s} and Eq. \ref{r_deltav}. In contrast, the speed reward function (Eq. \ref{r_v}) is only related to the speed of CV and is not influenced by this assumption. Therefore, the accuracies of the DDPGd and DDPG$\Delta$v models are lower than that of the DDPGv model.

\subsection{Evasive trajectory simulation}

\begin{figure}[!htpb]
    \centering
    \begin{subfigure}{0.9\linewidth}
        \centering
        \includegraphics[width=1\textwidth]{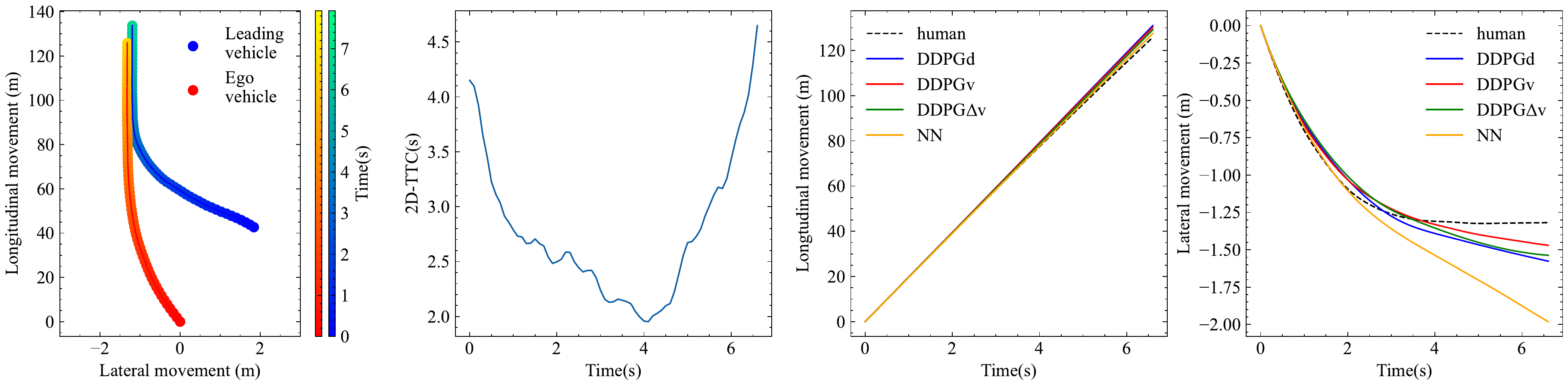}
        \caption{Device=10121, Trip=703, ObstacleId=39}\label{e1}
    \end{subfigure}
    
    \begin{subfigure}{0.9\linewidth}
        \centering
        \includegraphics[width=1\textwidth]{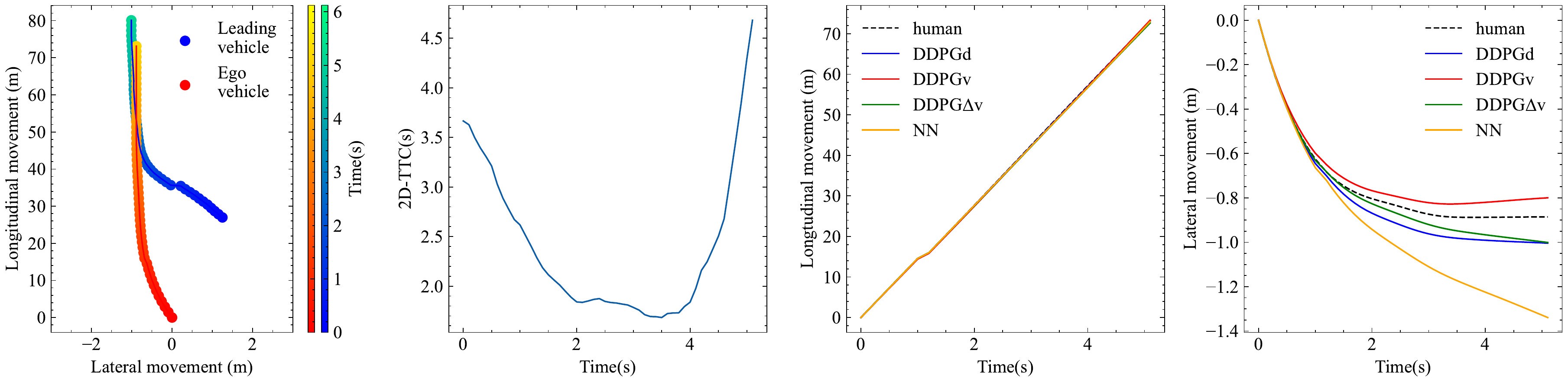}
        \caption{Device=10132, Trip=500, ObstacleId=29}\label{e2}
    \end{subfigure}
    
    \begin{subfigure}{0.9\linewidth}
        \centering
        \includegraphics[width=1\textwidth]{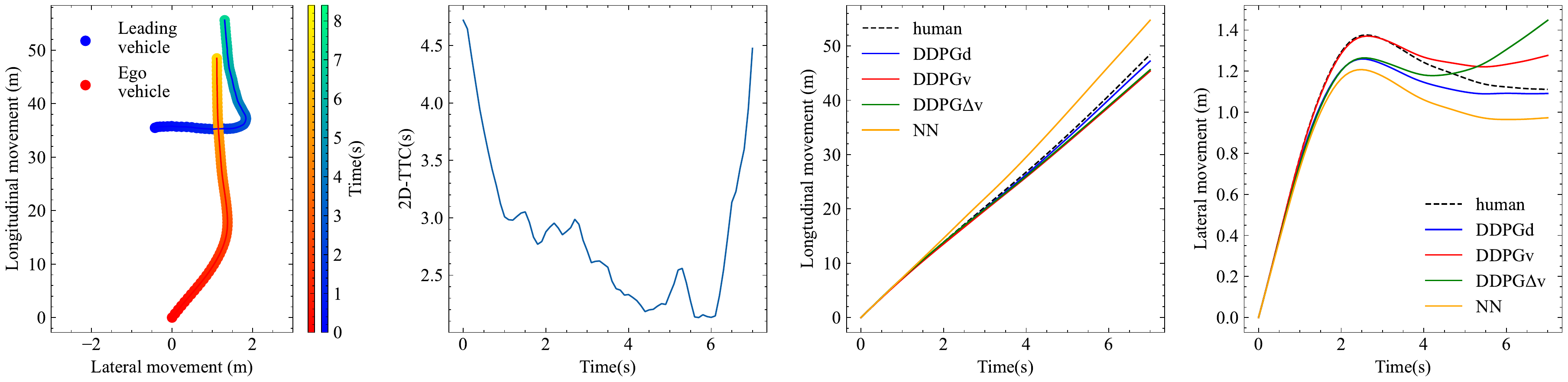}
        \caption{Device=17101, Trip=1039, ObstacleId=15}\label{e3}
    \end{subfigure}

    \begin{subfigure}{0.9\linewidth}
        \centering
        \includegraphics[width=1\textwidth]{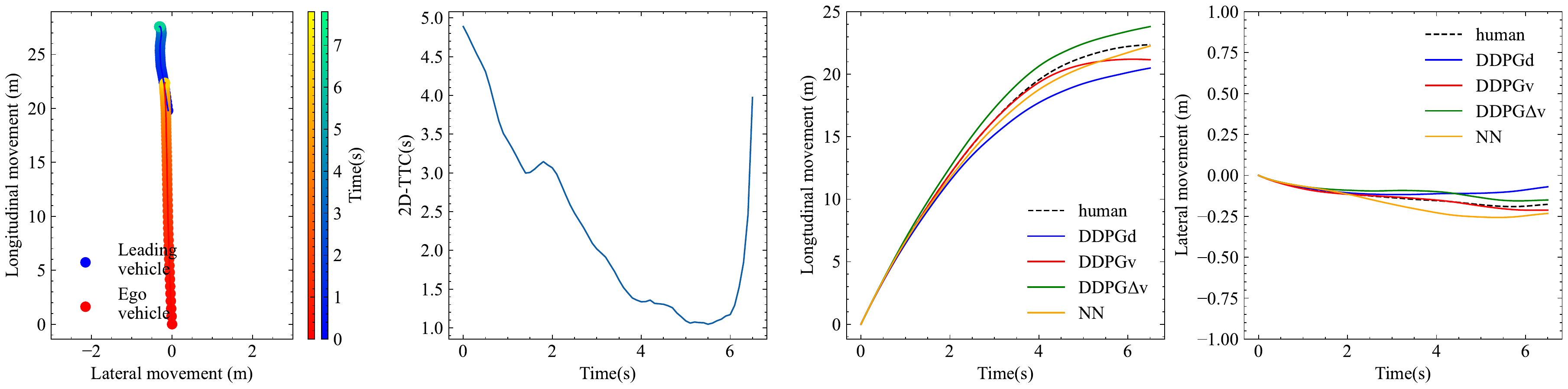}
        \caption{Device=10160, Trip=366, ObstacleId=15}\label{e4}
    \end{subfigure}
    
     \begin{subfigure}{0.9\linewidth}
        \centering
        \includegraphics[width=1\textwidth]{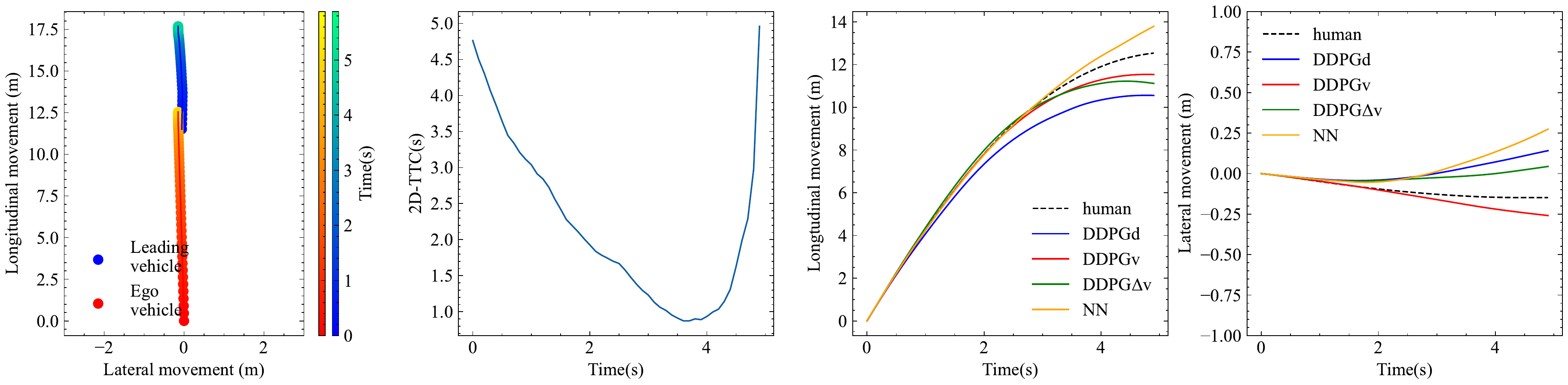}
        \caption{Device=10131, Trip=590, ObstacleId=36}\label{e5}
    \end{subfigure}
    \caption{Trajectory reconstruction examples}
    \label{trajectory_resconstruction}
\end{figure}

To demonstrate the implementation of the proposed 2D-TTC and DDPG approaches, several safety-critical scenarios were extracted and the evasive trajectories were simulated. Only the initial observed state was given to the models, and the whole trajectory was generated sequentially. The 2D-TTC values during the evasion process were also investigated. The original trajectories of the leading and ego vehicles, the 2D-TTC values, and the simulated longitudinal and lateral movements of the ego vehicle are illustrated in Fig. \ref{trajectory_resconstruction}. Specifically, Fig. \ref{e1}, \ref{e2}, and \ref{e3} demonstrate the situations where the ego vehicles avoid the conflict with cut-in vehicles by changing lanes, and Fig. \ref{e4} and \ref{e5} show the cases where the ego vehicles reduce speed as the leading vehicles decelerate.

Overall, the simulated trajectories are highly consistent with the human-driven ones, which indicates that the DDPG algorithms can model both the lane change and deceleration evasive behaviors accurately. The NN model performs worse than the DDPG ones, showing it fails to capture the inherent mechanism of evasive behaviors, and thus it is unable to process the unseen observations in the test set. Among the DDPG models, the DDPGv model predicts the trajectories closest to the observed ones in most cases. 

Besides, the 2D-TTC method can detect potential conflict in various situations precisely. The U-shaped 2D-TTC curve shows its ability to reflect the process of evasive behavior, from the danger realization to the successful collision avoidance. It is worth to mention that the 2D-TTC start from a value less than the set threshold in some cases, especially when the leading vehicle cuts in from another lane within a short distance. Because the view angle of the MobilEye main camera is $28^{\circ}$ \citep{stein2003vision}, the conflict cannot be captured until the leading vehicle enters the detection range. However, the trajectory suggests that the driver might realize the conflict before it is detected by the sensor and took evasive action.

\section{Conclusion}\label{conclusion}

In this study, DDPG-based driver's evasive behavior models during safety-critical lane changes were developed using the large-scale driving data collected by CVs. A novel 2D-TTC was proposed to capture the potential risks and identify safety-critical lane changes. It is capable to detect the risks in 2D traffic scenarios without the need for complete trajectories and is suitable for real-time risk estimation using the data collected by CVs. The correlation between the conflicts identified by 2D-TTC and the historical crashes was investigated. The results show that the risks captured by the 2D-TTC are highly correlated with the archived crashes. Furthermore, a DDPG-based evasive behavior model was developed based on the safety-critical situations detected by the 2D-TTC measure. The results indicate that the DDPG model can model both the longitudinal and lateral evasive behaviors accurately, and the simulated trajectories are highly consistent with the actual ones. The DDPG model also shows superior performance in understanding the conflict evasion strategies and predicting the precise trajectories compared to the conventional NN model. The evasive behavior model could facilitate the development of safety-aware microscopic simulations and predictive collision avoidance systems. For future studies, the transferability of the proposed methods will be further tested using the datasets in different driving environments.

\bibliography{mybibfile}

\begin{thebibliography}{106}
\expandafter\ifx\csname natexlab\endcsname\relax\def\natexlab#1{#1}\fi
\providecommand{\url}[1]{\texttt{#1}}
\providecommand{\href}[2]{#2}
\providecommand{\path}[1]{#1}
\providecommand{\DOIprefix}{doi:}
\providecommand{\ArXivprefix}{arXiv:}
\providecommand{\URLprefix}{URL: }
\providecommand{\Pubmedprefix}{pmid:}
\providecommand{\doi}[1]{\href{http://dx.doi.org/#1}{\path{#1}}}
\providecommand{\Pubmed}[1]{\href{pmid:#1}{\path{#1}}}
\providecommand{\bibinfo}[2]{#2}
\ifx\xfnm\relax \def\xfnm[#1]{\unskip,\space#1}\fi
\bibitem[{Abdulhai and Kattan(2003)}]{abdulhai2003reinforcement}
\bibinfo{author}{Abdulhai, B.}, \bibinfo{author}{Kattan, L.},
  \bibinfo{year}{2003}.
\newblock \bibinfo{title}{Reinforcement learning: Introduction to theory and
  potential for transport applications}.
\newblock \bibinfo{journal}{Canadian Journal of Civil Engineering}
  \bibinfo{volume}{30}, \bibinfo{pages}{981--991}.
\bibitem[{Ali et~al.(2020)Ali, Bliemer, Zheng and Haque}]{ali2020cooperate}
\bibinfo{author}{Ali, Y.}, \bibinfo{author}{Bliemer, M.C.},
  \bibinfo{author}{Zheng, Z.}, \bibinfo{author}{Haque, M.M.},
  \bibinfo{year}{2020}.
\newblock \bibinfo{title}{Cooperate or not? exploring drivers’ interactions
  and response times to a lane-changing request in a connected environment}.
\newblock \bibinfo{journal}{Transportation Research Part C: Emerging
  Technologies} \bibinfo{volume}{120}, \bibinfo{pages}{102816}.
\bibitem[{Alsaleh and Sayed(2021)}]{alsaleh2021markov}
\bibinfo{author}{Alsaleh, R.}, \bibinfo{author}{Sayed, T.},
  \bibinfo{year}{2021}.
\newblock \bibinfo{title}{Markov-game modeling of cyclist-pedestrian
  interactions in shared spaces: A multi-agent adversarial inverse
  reinforcement learning approach}.
\newblock \bibinfo{journal}{Transportation research part C: emerging
  technologies} \bibinfo{volume}{128}, \bibinfo{pages}{103191}.
\bibitem[{Aslani et~al.(2017)Aslani, Mesgari and Wiering}]{aslani2017adaptive}
\bibinfo{author}{Aslani, M.}, \bibinfo{author}{Mesgari, M.S.},
  \bibinfo{author}{Wiering, M.}, \bibinfo{year}{2017}.
\newblock \bibinfo{title}{Adaptive traffic signal control with actor-critic
  methods in a real-world traffic network with different traffic disruption
  events}.
\newblock \bibinfo{journal}{Transportation Research Part C: Emerging
  Technologies} \bibinfo{volume}{85}, \bibinfo{pages}{732--752}.
\bibitem[{Bezzina and Sayer(2014)}]{bezzina2014safety}
\bibinfo{author}{Bezzina, D.}, \bibinfo{author}{Sayer, J.},
  \bibinfo{year}{2014}.
\newblock \bibinfo{title}{Safety pilot model deployment: Test conductor team
  report}.
\newblock \bibinfo{type}{Technical Report} \bibinfo{number}{DOT HS 812 171}.
  National Highway Traffic Safety Administration.
\bibitem[{Biondi et~al.(2017)Biondi, Strayer, Rossi, Gastaldi and
  Mulatti}]{biondi2017advanced}
\bibinfo{author}{Biondi, F.}, \bibinfo{author}{Strayer, D.L.},
  \bibinfo{author}{Rossi, R.}, \bibinfo{author}{Gastaldi, M.},
  \bibinfo{author}{Mulatti, C.}, \bibinfo{year}{2017}.
\newblock \bibinfo{title}{Advanced driver assistance systems: Using multimodal
  redundant warnings to enhance road safety}.
\newblock \bibinfo{journal}{Applied ergonomics} \bibinfo{volume}{58},
  \bibinfo{pages}{238--244}.
\bibitem[{Chae et~al.(2017)Chae, Kang, Kim, Kim, Chung and
  Choi}]{chae2017autonomous}
\bibinfo{author}{Chae, H.}, \bibinfo{author}{Kang, C.M.}, \bibinfo{author}{Kim,
  B.}, \bibinfo{author}{Kim, J.}, \bibinfo{author}{Chung, C.C.},
  \bibinfo{author}{Choi, J.W.}, \bibinfo{year}{2017}.
\newblock \bibinfo{title}{Autonomous braking system via deep reinforcement
  learning}, in: \bibinfo{booktitle}{2017 IEEE 20th International conference on
  intelligent transportation systems (ITSC)}, \bibinfo{organization}{IEEE}. pp.
  \bibinfo{pages}{1--6}.
\bibitem[{Chen et~al.(2021)Chen, Huang, Li, Lee, Long, Gu and
  Zhai}]{chen2021modeling}
\bibinfo{author}{Chen, Q.}, \bibinfo{author}{Huang, H.}, \bibinfo{author}{Li,
  Y.}, \bibinfo{author}{Lee, J.}, \bibinfo{author}{Long, K.},
  \bibinfo{author}{Gu, R.}, \bibinfo{author}{Zhai, X.}, \bibinfo{year}{2021}.
\newblock \bibinfo{title}{Modeling accident risks in different lane-changing
  behavioral patterns}.
\newblock \bibinfo{journal}{Analytic methods in accident research}
  \bibinfo{volume}{30}, \bibinfo{pages}{100159}.
\bibitem[{Deveaux et~al.(2021)Deveaux, Higuchi, U{\c{c}}ar, Wang, H{\"a}rri and
  Altintas}]{deveaux2021extraction}
\bibinfo{author}{Deveaux, D.}, \bibinfo{author}{Higuchi, T.},
  \bibinfo{author}{U{\c{c}}ar, S.}, \bibinfo{author}{Wang, C.H.},
  \bibinfo{author}{H{\"a}rri, J.}, \bibinfo{author}{Altintas, O.},
  \bibinfo{year}{2021}.
\newblock \bibinfo{title}{Extraction of risk knowledge from time to collision
  variation in roundabouts}, in: \bibinfo{booktitle}{2021 IEEE International
  Intelligent Transportation Systems Conference (ITSC)},
  \bibinfo{organization}{IEEE}. pp. \bibinfo{pages}{3665--3672}.
\bibitem[{Dong et~al.(2021)Dong, Chen, Li, Du, Steinfeld and
  Labi}]{dong2021space}
\bibinfo{author}{Dong, J.}, \bibinfo{author}{Chen, S.}, \bibinfo{author}{Li,
  Y.}, \bibinfo{author}{Du, R.}, \bibinfo{author}{Steinfeld, A.},
  \bibinfo{author}{Labi, S.}, \bibinfo{year}{2021}.
\newblock \bibinfo{title}{Space-weighted information fusion using deep
  reinforcement learning: The context of tactical control of lane-changing
  autonomous vehicles and connectivity range assessment}.
\newblock \bibinfo{journal}{Transportation Research Part C: Emerging
  Technologies} \bibinfo{volume}{128}, \bibinfo{pages}{103192}.
\bibitem[{Dozza(2013)}]{dozza2013factors}
\bibinfo{author}{Dozza, M.}, \bibinfo{year}{2013}.
\newblock \bibinfo{title}{What factors influence drivers’ response time for
  evasive maneuvers in real traffic?}
\newblock \bibinfo{journal}{Accident Analysis \& Prevention}
  \bibinfo{volume}{58}, \bibinfo{pages}{299--308}.
\bibitem[{ESRI(2011)}]{arcgis}
\bibinfo{author}{ESRI}, \bibinfo{year}{2011}.
\newblock \bibinfo{title}{ArcGIS Desktop: Release 10}.
\newblock \bibinfo{publisher}{Environmental Systems Research Institute},
  \bibinfo{address}{Redlands, CA}.
\bibitem[{Essa and Sayed(2020)}]{essa2020self}
\bibinfo{author}{Essa, M.}, \bibinfo{author}{Sayed, T.}, \bibinfo{year}{2020}.
\newblock \bibinfo{title}{Self-learning adaptive traffic signal control for
  real-time safety optimization}.
\newblock \bibinfo{journal}{Accident Analysis \& Prevention}
  \bibinfo{volume}{146}, \bibinfo{pages}{105713}.
\bibitem[{Farazi et~al.(2021)Farazi, Zou, Ahamed and Barua}]{farazi2021deep}
\bibinfo{author}{Farazi, N.P.}, \bibinfo{author}{Zou, B.},
  \bibinfo{author}{Ahamed, T.}, \bibinfo{author}{Barua, L.},
  \bibinfo{year}{2021}.
\newblock \bibinfo{title}{Deep reinforcement learning in transportation
  research: A review}.
\newblock \bibinfo{journal}{Transportation research interdisciplinary
  perspectives} \bibinfo{volume}{11}, \bibinfo{pages}{100425}.
\bibitem[{Gettman and Head(2003)}]{gettman2003surrogate}
\bibinfo{author}{Gettman, D.}, \bibinfo{author}{Head, L.},
  \bibinfo{year}{2003}.
\newblock \bibinfo{title}{Surrogate safety measures from traffic simulation
  models}.
\newblock \bibinfo{journal}{Transportation Research Record}
  \bibinfo{volume}{1840}, \bibinfo{pages}{104--115}.
\bibitem[{Guo et~al.(2022)Guo, Keyvan-Ekbatani and Xie}]{guo2022lane}
\bibinfo{author}{Guo, H.}, \bibinfo{author}{Keyvan-Ekbatani, M.},
  \bibinfo{author}{Xie, K.}, \bibinfo{year}{2022}.
\newblock \bibinfo{title}{Lane change detection and prediction using real-world
  connected vehicle data}.
\newblock \bibinfo{journal}{Transportation Research Part C: Emerging
  Technologies} \bibinfo{volume}{142}, \bibinfo{pages}{103785}.
\bibitem[{Guo et~al.(2021)Guo, Xie and Keyvan-Ekbatani}]{guo2021lane}
\bibinfo{author}{Guo, H.}, \bibinfo{author}{Xie, K.},
  \bibinfo{author}{Keyvan-Ekbatani, M.}, \bibinfo{year}{2021}.
\newblock \bibinfo{title}{Lane change detection using naturalistic driving
  data}, in: \bibinfo{booktitle}{2021 7th International Conference on Models
  and Technologies for Intelligent Transportation Systems (MT-ITS)},
  \bibinfo{organization}{IEEE}. pp. \bibinfo{pages}{1--6}.
\bibitem[{Hamilton and Allen(2015)}]{hamilton2015safety}
\bibinfo{author}{Hamilton, B.}, \bibinfo{author}{Allen}, \bibinfo{year}{2015}.
\newblock \bibinfo{title}{Safety pilot model deployment-sample data environment
  data handbook}.
\newblock \bibinfo{type}{Technical Report}. US department of transportation.
\bibitem[{Happee et~al.(2017)Happee, Gold, Radlmayr, Hergeth and
  Bengler}]{happee2017take}
\bibinfo{author}{Happee, R.}, \bibinfo{author}{Gold, C.},
  \bibinfo{author}{Radlmayr, J.}, \bibinfo{author}{Hergeth, S.},
  \bibinfo{author}{Bengler, K.}, \bibinfo{year}{2017}.
\newblock \bibinfo{title}{Take-over performance in evasive manoeuvres}.
\newblock \bibinfo{journal}{Accident Analysis \& Prevention}
  \bibinfo{volume}{106}, \bibinfo{pages}{211--222}.
\bibitem[{Harding et~al.(2014)Harding, Powell, Yoon, Fikentscher, Doyle, Sade,
  Lukuc, Simons, Wang et~al.}]{harding2014vehicle}
\bibinfo{author}{Harding, J.}, \bibinfo{author}{Powell, G.},
  \bibinfo{author}{Yoon, R.}, \bibinfo{author}{Fikentscher, J.},
  \bibinfo{author}{Doyle, C.}, \bibinfo{author}{Sade, D.},
  \bibinfo{author}{Lukuc, M.}, \bibinfo{author}{Simons, J.},
  \bibinfo{author}{Wang, J.}, et~al., \bibinfo{year}{2014}.
\newblock \bibinfo{title}{Vehicle-to-vehicle communications: readiness of V2V
  technology for application}.
\newblock \bibinfo{type}{Technical Report} \bibinfo{number}{DOT HS 812 014}.
  National Highway Traffic Safety Administration.
\bibitem[{Haydari and Yilmaz(2020)}]{haydari2020deep}
\bibinfo{author}{Haydari, A.}, \bibinfo{author}{Yilmaz, Y.},
  \bibinfo{year}{2020}.
\newblock \bibinfo{title}{Deep reinforcement learning for intelligent
  transportation systems: A survey}.
\newblock \bibinfo{journal}{IEEE Transactions on Intelligent Transportation
  Systems} .
\bibitem[{Hayward(1972)}]{hayward1972near}
\bibinfo{author}{Hayward, J.C.}, \bibinfo{year}{1972}.
\newblock \bibinfo{title}{Near miss determination through use of a scale of
  danger}.
\newblock \bibinfo{journal}{Transportation Research Record} ,
  \bibinfo{pages}{24--34}.
\bibitem[{Henclewood et~al.(2014)Henclewood, Abramovich and
  Yelchuru}]{henclewood2014safety}
\bibinfo{author}{Henclewood, D.}, \bibinfo{author}{Abramovich, M.},
  \bibinfo{author}{Yelchuru, B.}, \bibinfo{year}{2014}.
\newblock \bibinfo{title}{Safety pilot model deployment--one day sample data
  environment data handbook}.
\newblock \bibinfo{type}{Technical Report}. Research and Technology Innovation
  Administration, US Department of Transportation.
\bibitem[{Horiuchi et~al.(2001)Horiuchi, Okada and
  Nohtomi}]{horiuchi2001numerical}
\bibinfo{author}{Horiuchi, S.}, \bibinfo{author}{Okada, K.},
  \bibinfo{author}{Nohtomi, S.}, \bibinfo{year}{2001}.
\newblock \bibinfo{title}{Numerical analysis of optimal vehicle trajectories
  for emergency obstacle avoidance}.
\newblock \bibinfo{journal}{JSAE review} \bibinfo{volume}{22},
  \bibinfo{pages}{495--502}.
\bibitem[{Hou et~al.(2014)Hou, List and Guo}]{hou2014new}
\bibinfo{author}{Hou, J.}, \bibinfo{author}{List, G.F.}, \bibinfo{author}{Guo,
  X.}, \bibinfo{year}{2014}.
\newblock \bibinfo{title}{New algorithms for computing the time-to-collision in
  freeway traffic simulation models}.
\newblock \bibinfo{journal}{Computational intelligence and neuroscience}
  \bibinfo{volume}{2014}.
\bibitem[{Huang et~al.(2018)Huang, Sun and Sun}]{huang2018car}
\bibinfo{author}{Huang, X.}, \bibinfo{author}{Sun, J.}, \bibinfo{author}{Sun,
  J.}, \bibinfo{year}{2018}.
\newblock \bibinfo{title}{A car-following model considering asymmetric driving
  behavior based on long short-term memory neural networks}.
\newblock \bibinfo{journal}{Transportation research part C: emerging
  technologies} \bibinfo{volume}{95}, \bibinfo{pages}{346--362}.
\bibitem[{Huang et~al.(2017)Huang, Zhao and Peng}]{huang2017empirical}
\bibinfo{author}{Huang, X.}, \bibinfo{author}{Zhao, D.}, \bibinfo{author}{Peng,
  H.}, \bibinfo{year}{2017}.
\newblock \bibinfo{title}{Empirical study of dsrc performance based on safety
  pilot model deployment data}.
\newblock \bibinfo{journal}{IEEE Transactions on Intelligent Transportation
  Systems} \bibinfo{volume}{18}, \bibinfo{pages}{2619--2628}.
\bibitem[{Jamson et~al.(2013)Jamson, Merat, Carsten and
  Lai}]{jamson2013behavioural}
\bibinfo{author}{Jamson, A.H.}, \bibinfo{author}{Merat, N.},
  \bibinfo{author}{Carsten, O.M.}, \bibinfo{author}{Lai, F.C.},
  \bibinfo{year}{2013}.
\newblock \bibinfo{title}{Behavioural changes in drivers experiencing
  highly-automated vehicle control in varying traffic conditions}.
\newblock \bibinfo{journal}{Transportation research part C: emerging
  technologies} \bibinfo{volume}{30}, \bibinfo{pages}{116--125}.
\bibitem[{Jurecki and Sta{\'n}czyk(2009)}]{jurecki2009driver}
\bibinfo{author}{Jurecki, R.}, \bibinfo{author}{Sta{\'n}czyk, T.},
  \bibinfo{year}{2009}.
\newblock \bibinfo{title}{Driver model for the analysis of pre-accident
  situations}.
\newblock \bibinfo{journal}{Vehicle System Dynamics} \bibinfo{volume}{47},
  \bibinfo{pages}{589--612}.
\bibitem[{{Kelly Blue Book}(2013)}]{kbb2013}
\bibinfo{author}{{Kelly Blue Book}}, \bibinfo{year}{2013}.
\newblock \bibinfo{title}{Highest horsepower sedans of 2013}.
\newblock
  \bibinfo{howpublished}{\url{https://www.kbb.com/highest-horsepower-cars/sedan/2013/}}.
\bibitem[{Keyvan-Ekbatani et~al.(2016)Keyvan-Ekbatani, Knoop and
  Daamen}]{keyvan2016categorization}
\bibinfo{author}{Keyvan-Ekbatani, M.}, \bibinfo{author}{Knoop, V.L.},
  \bibinfo{author}{Daamen, W.}, \bibinfo{year}{2016}.
\newblock \bibinfo{title}{Categorization of the lane change decision process on
  freeways}.
\newblock \bibinfo{journal}{Transportation research part C: emerging
  technologies} \bibinfo{volume}{69}, \bibinfo{pages}{515--526}.
\bibitem[{Kuang et~al.(2015)Kuang, Qu and Wang}]{kuang2015tree}
\bibinfo{author}{Kuang, Y.}, \bibinfo{author}{Qu, X.}, \bibinfo{author}{Wang,
  S.}, \bibinfo{year}{2015}.
\newblock \bibinfo{title}{A tree-structured crash surrogate measure for
  freeways}.
\newblock \bibinfo{journal}{Accident Analysis \& Prevention}
  \bibinfo{volume}{77}, \bibinfo{pages}{137--148}.
\bibitem[{Lee et~al.(2019)Lee, Ngoduy and Keyvan-Ekbatani}]{lee2019integrated}
\bibinfo{author}{Lee, S.}, \bibinfo{author}{Ngoduy, D.},
  \bibinfo{author}{Keyvan-Ekbatani, M.}, \bibinfo{year}{2019}.
\newblock \bibinfo{title}{Integrated deep learning and stochastic car-following
  model for traffic dynamics on multi-lane freeways}.
\newblock \bibinfo{journal}{Transportation research part C: emerging
  technologies} \bibinfo{volume}{106}, \bibinfo{pages}{360--377}.
\bibitem[{Lenard et~al.(2018)Lenard, Welsh and Danton}]{lenard2018time}
\bibinfo{author}{Lenard, J.}, \bibinfo{author}{Welsh, R.},
  \bibinfo{author}{Danton, R.}, \bibinfo{year}{2018}.
\newblock \bibinfo{title}{Time-to-collision analysis of pedestrian and
  pedal-cycle accidents for the development of autonomous emergency braking
  systems}.
\newblock \bibinfo{journal}{Accident Analysis \& Prevention}
  \bibinfo{volume}{115}, \bibinfo{pages}{128--136}.
\bibitem[{Li et~al.(2022)Li, Yang, Li, Qu, Lyu and Li}]{li2022decision}
\bibinfo{author}{Li, G.}, \bibinfo{author}{Yang, Y.}, \bibinfo{author}{Li, S.},
  \bibinfo{author}{Qu, X.}, \bibinfo{author}{Lyu, N.}, \bibinfo{author}{Li,
  S.E.}, \bibinfo{year}{2022}.
\newblock \bibinfo{title}{Decision making of autonomous vehicles in lane change
  scenarios: Deep reinforcement learning approaches with risk awareness}.
\newblock \bibinfo{journal}{Transportation research part C: emerging
  technologies} \bibinfo{volume}{134}, \bibinfo{pages}{103452}.
\bibitem[{Li(2017)}]{li2017deep}
\bibinfo{author}{Li, Y.}, \bibinfo{year}{2017}.
\newblock \bibinfo{title}{Deep reinforcement learning: An overview}.
\newblock \bibinfo{journal}{arXiv preprint arXiv:1701.07274} .
\bibitem[{Li et~al.(2017)Li, Wang, Wang, Xing, Liu and Wei}]{li2017evaluation}
\bibinfo{author}{Li, Y.}, \bibinfo{author}{Wang, H.}, \bibinfo{author}{Wang,
  W.}, \bibinfo{author}{Xing, L.}, \bibinfo{author}{Liu, S.},
  \bibinfo{author}{Wei, X.}, \bibinfo{year}{2017}.
\newblock \bibinfo{title}{Evaluation of the impacts of cooperative adaptive
  cruise control on reducing rear-end collision risks on freeways}.
\newblock \bibinfo{journal}{Accident Analysis \& Prevention}
  \bibinfo{volume}{98}, \bibinfo{pages}{87--95}.
\bibitem[{Li et~al.(2021)Li, Wu, Chen, Lee and Long}]{li2021exploring}
\bibinfo{author}{Li, Y.}, \bibinfo{author}{Wu, D.}, \bibinfo{author}{Chen, Q.},
  \bibinfo{author}{Lee, J.}, \bibinfo{author}{Long, K.}, \bibinfo{year}{2021}.
\newblock \bibinfo{title}{Exploring transition durations of rear-end collisions
  based on vehicle trajectory data: a survival modeling approach}.
\newblock \bibinfo{journal}{Accident Analysis \& Prevention}
  \bibinfo{volume}{159}, \bibinfo{pages}{106271}.
\bibitem[{Li et~al.(2020)Li, Wu, Lee, Yang and Shi}]{li2020analysis}
\bibinfo{author}{Li, Y.}, \bibinfo{author}{Wu, D.}, \bibinfo{author}{Lee, J.},
  \bibinfo{author}{Yang, M.}, \bibinfo{author}{Shi, Y.}, \bibinfo{year}{2020}.
\newblock \bibinfo{title}{Analysis of the transition condition of rear-end
  collisions using time-to-collision index and vehicle trajectory data}.
\newblock \bibinfo{journal}{Accident Analysis \& Prevention}
  \bibinfo{volume}{144}, \bibinfo{pages}{105676}.
\bibitem[{Lillicrap et~al.(2015)Lillicrap, Hunt, Pritzel, Heess, Erez, Tassa,
  Silver and Wierstra}]{lillicrap2015continuous}
\bibinfo{author}{Lillicrap, T.P.}, \bibinfo{author}{Hunt, J.J.},
  \bibinfo{author}{Pritzel, A.}, \bibinfo{author}{Heess, N.},
  \bibinfo{author}{Erez, T.}, \bibinfo{author}{Tassa, Y.},
  \bibinfo{author}{Silver, D.}, \bibinfo{author}{Wierstra, D.},
  \bibinfo{year}{2015}.
\newblock \bibinfo{title}{Continuous control with deep reinforcement learning}.
\newblock \bibinfo{journal}{arXiv preprint arXiv:1509.02971} .
\bibitem[{Ma and Qu(2020)}]{ma2020sequence}
\bibinfo{author}{Ma, L.}, \bibinfo{author}{Qu, S.}, \bibinfo{year}{2020}.
\newblock \bibinfo{title}{A sequence to sequence learning based car-following
  model for multi-step predictions considering reaction delay}.
\newblock \bibinfo{journal}{Transportation research part C: emerging
  technologies} \bibinfo{volume}{120}, \bibinfo{pages}{102785}.
\bibitem[{Ma and Ahn(2008)}]{ma2008comparisons}
\bibinfo{author}{Ma, T.}, \bibinfo{author}{Ahn, S.}, \bibinfo{year}{2008}.
\newblock \bibinfo{title}{Comparisons of speed-spacing relations under general
  car following versus lane changing}.
\newblock \bibinfo{journal}{Transportation research record}
  \bibinfo{volume}{2088}, \bibinfo{pages}{138--147}.
\bibitem[{Markkula(2014)}]{markkula2014modeling}
\bibinfo{author}{Markkula, G.}, \bibinfo{year}{2014}.
\newblock \bibinfo{title}{Modeling driver control behavior in both routine and
  near-accident driving}, in: \bibinfo{booktitle}{Proceedings of the human
  factors and ergonomics society annual meeting}, \bibinfo{organization}{SAGE
  Publications Sage CA: Los Angeles, CA}. pp. \bibinfo{pages}{879--883}.
\bibitem[{Markkula et~al.(2012)Markkula, Benderius, Wolff and
  Wahde}]{markkula2012review}
\bibinfo{author}{Markkula, G.}, \bibinfo{author}{Benderius, O.},
  \bibinfo{author}{Wolff, K.}, \bibinfo{author}{Wahde, M.},
  \bibinfo{year}{2012}.
\newblock \bibinfo{title}{A review of near-collision driver behavior models}.
\newblock \bibinfo{journal}{Human factors} \bibinfo{volume}{54},
  \bibinfo{pages}{1117--1143}.
\bibitem[{Markkula et~al.(2016)Markkula, Engstr{\"o}m, Lodin, B{\"a}rgman and
  Victor}]{markkula2016farewell}
\bibinfo{author}{Markkula, G.}, \bibinfo{author}{Engstr{\"o}m, J.},
  \bibinfo{author}{Lodin, J.}, \bibinfo{author}{B{\"a}rgman, J.},
  \bibinfo{author}{Victor, T.}, \bibinfo{year}{2016}.
\newblock \bibinfo{title}{A farewell to brake reaction times?
  kinematics-dependent brake response in naturalistic rear-end emergencies}.
\newblock \bibinfo{journal}{Accident Analysis \& Prevention}
  \bibinfo{volume}{95}, \bibinfo{pages}{209--226}.
\bibitem[{McGehee et~al.(1999)McGehee, Mazzae, Baldwin, Grant, Simmons, Hankey
  and Forkenbrock}]{mcgehee1999examination}
\bibinfo{author}{McGehee, D.V.}, \bibinfo{author}{Mazzae, E.N.},
  \bibinfo{author}{Baldwin, G.S.}, \bibinfo{author}{Grant, P.},
  \bibinfo{author}{Simmons, C.J.}, \bibinfo{author}{Hankey, J.},
  \bibinfo{author}{Forkenbrock, G.}, \bibinfo{year}{1999}.
\newblock \bibinfo{title}{Examination of drivers' collision avoidance behavior
  using conventional and antilock brake systems on the iowa driving simulator}.
\newblock \bibinfo{type}{Technical Report}. University of Iowa.
\bibitem[{Minderhoud and Bovy(2001)}]{minderhoud2001extended}
\bibinfo{author}{Minderhoud, M.M.}, \bibinfo{author}{Bovy, P.H.},
  \bibinfo{year}{2001}.
\newblock \bibinfo{title}{Extended time-to-collision measures for road traffic
  safety assessment}.
\newblock \bibinfo{journal}{Accident Analysis \& Prevention}
  \bibinfo{volume}{33}, \bibinfo{pages}{89--97}.
\bibitem[{Mnih et~al.(2015)Mnih, Kavukcuoglu, Silver, Rusu, Veness, Bellemare,
  Graves, Riedmiller, Fidjeland, Ostrovski et~al.}]{mnih2015human}
\bibinfo{author}{Mnih, V.}, \bibinfo{author}{Kavukcuoglu, K.},
  \bibinfo{author}{Silver, D.}, \bibinfo{author}{Rusu, A.A.},
  \bibinfo{author}{Veness, J.}, \bibinfo{author}{Bellemare, M.G.},
  \bibinfo{author}{Graves, A.}, \bibinfo{author}{Riedmiller, M.},
  \bibinfo{author}{Fidjeland, A.K.}, \bibinfo{author}{Ostrovski, G.}, et~al.,
  \bibinfo{year}{2015}.
\newblock \bibinfo{title}{Human-level control through deep reinforcement
  learning}.
\newblock \bibinfo{journal}{nature} \bibinfo{volume}{518},
  \bibinfo{pages}{529--533}.
\bibitem[{Mohebbi et~al.(2009)Mohebbi, Gray and Tan}]{mohebbi2009driver}
\bibinfo{author}{Mohebbi, R.}, \bibinfo{author}{Gray, R.},
  \bibinfo{author}{Tan, H.Z.}, \bibinfo{year}{2009}.
\newblock \bibinfo{title}{Driver reaction time to tactile and auditory rear-end
  collision warnings while talking on a cell phone}.
\newblock \bibinfo{journal}{Human factors} \bibinfo{volume}{51},
  \bibinfo{pages}{102--110}.
\bibitem[{Najm et~al.(2013)Najm, Ranganathan, Srinivasan, Smith, Toma, Swanson,
  Burgett et~al.}]{najm2013description}
\bibinfo{author}{Najm, W.G.}, \bibinfo{author}{Ranganathan, R.},
  \bibinfo{author}{Srinivasan, G.}, \bibinfo{author}{Smith, J.D.},
  \bibinfo{author}{Toma, S.}, \bibinfo{author}{Swanson, E.},
  \bibinfo{author}{Burgett, A.}, et~al., \bibinfo{year}{2013}.
\newblock \bibinfo{title}{Description of light-vehicle pre-crash scenarios for
  safety applications based on vehicle-to-vehicle communications}.
\newblock \bibinfo{type}{Technical Report}. United States. National Highway
  Traffic Safety Administration.
\bibitem[{Nasernejad et~al.(2021)Nasernejad, Sayed and
  Alsaleh}]{nasernejad2021modeling}
\bibinfo{author}{Nasernejad, P.}, \bibinfo{author}{Sayed, T.},
  \bibinfo{author}{Alsaleh, R.}, \bibinfo{year}{2021}.
\newblock \bibinfo{title}{Modeling pedestrian behavior in pedestrian-vehicle
  near misses: A continuous gaussian process inverse reinforcement learning
  (gp-irl) approach}.
\newblock \bibinfo{journal}{Accident Analysis \& Prevention}
  \bibinfo{volume}{161}, \bibinfo{pages}{106355}.
\bibitem[{Ozan et~al.(2015)Ozan, Baskan, Haldenbilen and
  Ceylan}]{ozan2015modified}
\bibinfo{author}{Ozan, C.}, \bibinfo{author}{Baskan, O.},
  \bibinfo{author}{Haldenbilen, S.}, \bibinfo{author}{Ceylan, H.},
  \bibinfo{year}{2015}.
\newblock \bibinfo{title}{A modified reinforcement learning algorithm for
  solving coordinated signalized networks}.
\newblock \bibinfo{journal}{Transportation Research Part C: Emerging
  Technologies} \bibinfo{volume}{54}, \bibinfo{pages}{40--55}.
\bibitem[{Ozbay et~al.(2008)Ozbay, Yang, Bartin and
  Mudigonda}]{ozbay2008derivation}
\bibinfo{author}{Ozbay, K.}, \bibinfo{author}{Yang, H.},
  \bibinfo{author}{Bartin, B.}, \bibinfo{author}{Mudigonda, S.},
  \bibinfo{year}{2008}.
\newblock \bibinfo{title}{Derivation and validation of new simulation-based
  surrogate safety measure}.
\newblock \bibinfo{journal}{Transportation research record}
  \bibinfo{volume}{2083}, \bibinfo{pages}{105--113}.
\bibitem[{Papini et~al.(2021)Papini, Plebe, Da~Lio and
  Don{\`a}}]{papini2021reinforcement}
\bibinfo{author}{Papini, G.P.R.}, \bibinfo{author}{Plebe, A.},
  \bibinfo{author}{Da~Lio, M.}, \bibinfo{author}{Don{\`a}, R.},
  \bibinfo{year}{2021}.
\newblock \bibinfo{title}{A reinforcement learning approach for enacting
  cautious behaviours in autonomous driving system: Safe speed choice in the
  interaction with distracted pedestrians}.
\newblock \bibinfo{journal}{IEEE Transactions on Intelligent Transportation
  Systems} .
\bibitem[{Pearson(1895)}]{pearson1895vii}
\bibinfo{author}{Pearson, K.}, \bibinfo{year}{1895}.
\newblock \bibinfo{title}{Vii. note on regression and inheritance in the case
  of two parents}.
\newblock \bibinfo{journal}{proceedings of the royal society of London}
  \bibinfo{volume}{58}, \bibinfo{pages}{240--242}.
\bibitem[{Petersen and Barrett(2009)}]{petersen2009postural}
\bibinfo{author}{Petersen, A.}, \bibinfo{author}{Barrett, R.},
  \bibinfo{year}{2009}.
\newblock \bibinfo{title}{Postural stability and vehicle kinematics during an
  evasive lane change manoeuvre: A driver training study}.
\newblock \bibinfo{journal}{Ergonomics} \bibinfo{volume}{52},
  \bibinfo{pages}{560--568}.
\bibitem[{Pinnow et~al.(2021)Pinnow, Masoud, Elhenawy and
  Glaser}]{pinnow2021review}
\bibinfo{author}{Pinnow, J.}, \bibinfo{author}{Masoud, M.},
  \bibinfo{author}{Elhenawy, M.}, \bibinfo{author}{Glaser, S.},
  \bibinfo{year}{2021}.
\newblock \bibinfo{title}{A review of naturalistic driving study surrogates and
  surrogate indicator viability within the context of different road
  geometries}.
\newblock \bibinfo{journal}{Accident Analysis \& Prevention}
  \bibinfo{volume}{157}, \bibinfo{pages}{106185}.
\bibitem[{Sarkar et~al.(2021)Sarkar, Hickman, McDonald, Huang, Vogelpohl and
  Markkula}]{sarkar2021steering}
\bibinfo{author}{Sarkar, A.}, \bibinfo{author}{Hickman, J.S.},
  \bibinfo{author}{McDonald, A.D.}, \bibinfo{author}{Huang, W.},
  \bibinfo{author}{Vogelpohl, T.}, \bibinfo{author}{Markkula, G.},
  \bibinfo{year}{2021}.
\newblock \bibinfo{title}{Steering or braking avoidance response in shrp2
  rear-end crashes and near-crashes: A decision tree approach}.
\newblock \bibinfo{journal}{Accident Analysis \& Prevention}
  \bibinfo{volume}{154}, \bibinfo{pages}{106055}.
\bibitem[{Scanlon et~al.(2021)Scanlon, Kusano, Daniel, Alderson, Ogle and
  Victor}]{scanlon2021waymo}
\bibinfo{author}{Scanlon, J.M.}, \bibinfo{author}{Kusano, K.D.},
  \bibinfo{author}{Daniel, T.}, \bibinfo{author}{Alderson, C.},
  \bibinfo{author}{Ogle, A.}, \bibinfo{author}{Victor, T.},
  \bibinfo{year}{2021}.
\newblock \bibinfo{title}{Waymo simulated driving behavior in reconstructed
  fatal crashes within an autonomous vehicle operating domain}.
\newblock \bibinfo{journal}{Accident Analysis \& Prevention}
  \bibinfo{volume}{163}, \bibinfo{pages}{106454}.
\bibitem[{Scanlon et~al.(2015)Scanlon, Kusano and Gabler}]{scanlon2015analysis}
\bibinfo{author}{Scanlon, J.M.}, \bibinfo{author}{Kusano, K.D.},
  \bibinfo{author}{Gabler, H.C.}, \bibinfo{year}{2015}.
\newblock \bibinfo{title}{Analysis of driver evasive maneuvering prior to
  intersection crashes using event data recorders}.
\newblock \bibinfo{journal}{Traffic injury prevention} \bibinfo{volume}{16},
  \bibinfo{pages}{S182--S189}.
\bibitem[{Schmidt et~al.(2006)Schmidt, Oechsle and Branz}]{schmidt2006research}
\bibinfo{author}{Schmidt, C.}, \bibinfo{author}{Oechsle, F.},
  \bibinfo{author}{Branz, W.}, \bibinfo{year}{2006}.
\newblock \bibinfo{title}{Research on trajectory planning in emergency
  situations with multiple objects}, in: \bibinfo{booktitle}{2006 IEEE
  Intelligent Transportation Systems Conference}, \bibinfo{organization}{IEEE}.
  pp. \bibinfo{pages}{988--992}.
\bibitem[{Schnelle et~al.(2018)Schnelle, Wang, Jagacinski and
  Su}]{schnelle2018feedforward}
\bibinfo{author}{Schnelle, S.}, \bibinfo{author}{Wang, J.},
  \bibinfo{author}{Jagacinski, R.}, \bibinfo{author}{Su, H.j.},
  \bibinfo{year}{2018}.
\newblock \bibinfo{title}{A feedforward and feedback integrated lateral and
  longitudinal driver model for personalized advanced driver assistance
  systems}.
\newblock \bibinfo{journal}{Mechatronics} \bibinfo{volume}{50},
  \bibinfo{pages}{177--188}.
\bibitem[{Scott and Gray(2008)}]{scott2008comparison}
\bibinfo{author}{Scott, J.}, \bibinfo{author}{Gray, R.}, \bibinfo{year}{2008}.
\newblock \bibinfo{title}{A comparison of tactile, visual, and auditory
  warnings for rear-end collision prevention in simulated driving}.
\newblock \bibinfo{journal}{Human factors} \bibinfo{volume}{50},
  \bibinfo{pages}{264--275}.
\bibitem[{Sen et~al.(2003)Sen, Smith, Najm et~al.}]{sen2003analysis}
\bibinfo{author}{Sen, B.}, \bibinfo{author}{Smith, J.D.},
  \bibinfo{author}{Najm, W.G.}, et~al., \bibinfo{year}{2003}.
\newblock \bibinfo{title}{Analysis of lane change crashes}.
\newblock \bibinfo{type}{Technical Report}. United States. National Highway
  Traffic Safety Administration.
\bibitem[{Shi et~al.(2021)Shi, Zhou, Wu, Wang, Lin and Ran}]{shi2021connected}
\bibinfo{author}{Shi, H.}, \bibinfo{author}{Zhou, Y.}, \bibinfo{author}{Wu,
  K.}, \bibinfo{author}{Wang, X.}, \bibinfo{author}{Lin, Y.},
  \bibinfo{author}{Ran, B.}, \bibinfo{year}{2021}.
\newblock \bibinfo{title}{Connected automated vehicle cooperative control with
  a deep reinforcement learning approach in a mixed traffic environment}.
\newblock \bibinfo{journal}{Transportation Research Part C: Emerging
  Technologies} \bibinfo{volume}{133}, \bibinfo{pages}{103421}.
\bibitem[{Shibata et~al.(2014)Shibata, Sugiyama and
  Wada}]{shibata2014collision}
\bibinfo{author}{Shibata, N.}, \bibinfo{author}{Sugiyama, S.},
  \bibinfo{author}{Wada, T.}, \bibinfo{year}{2014}.
\newblock \bibinfo{title}{Collision avoidance control with steering using
  velocity potential field}, in: \bibinfo{booktitle}{2014 IEEE Intelligent
  Vehicles Symposium Proceedings}, \bibinfo{organization}{IEEE}. pp.
  \bibinfo{pages}{438--443}.
\bibitem[{Silver et~al.(2014)Silver, Lever, Heess, Degris, Wierstra and
  Riedmiller}]{silver2014deterministic}
\bibinfo{author}{Silver, D.}, \bibinfo{author}{Lever, G.},
  \bibinfo{author}{Heess, N.}, \bibinfo{author}{Degris, T.},
  \bibinfo{author}{Wierstra, D.}, \bibinfo{author}{Riedmiller, M.},
  \bibinfo{year}{2014}.
\newblock \bibinfo{title}{Deterministic policy gradient algorithms}, in:
  \bibinfo{booktitle}{International conference on machine learning},
  \bibinfo{organization}{PMLR}. pp. \bibinfo{pages}{387--395}.
\bibitem[{Soudbakhsh et~al.(2011)Soudbakhsh, Eskandarian and
  Moreau}]{soudbakhsh2011emergency}
\bibinfo{author}{Soudbakhsh, D.}, \bibinfo{author}{Eskandarian, A.},
  \bibinfo{author}{Moreau, J.}, \bibinfo{year}{2011}.
\newblock \bibinfo{title}{An emergency evasive maneuver algorithm for
  vehicles}, in: \bibinfo{booktitle}{2011 14th International IEEE Conference on
  Intelligent Transportation Systems (ITSC)}, \bibinfo{organization}{IEEE}. pp.
  \bibinfo{pages}{973--978}.
\bibitem[{Stein et~al.(2003)Stein, Mano and Shashua}]{stein2003vision}
\bibinfo{author}{Stein, G.P.}, \bibinfo{author}{Mano, O.},
  \bibinfo{author}{Shashua, A.}, \bibinfo{year}{2003}.
\newblock \bibinfo{title}{Vision-based acc with a single camera: bounds on
  range and range rate accuracy}, in: \bibinfo{booktitle}{IEEE IV2003
  intelligent vehicles symposium. Proceedings (Cat. No. 03TH8683)},
  \bibinfo{organization}{IEEE}. pp. \bibinfo{pages}{120--125}.
\bibitem[{Sugimoto and Sauer(2005)}]{sugimoto2005effectiveness}
\bibinfo{author}{Sugimoto, Y.}, \bibinfo{author}{Sauer, C.},
  \bibinfo{year}{2005}.
\newblock \bibinfo{title}{Effectiveness estimation method for advanced driver
  assistance system and its application to collision mitigation brake system},
  in: \bibinfo{booktitle}{Proceedings of the 19th International Technical
  Conference on the Enhanced Safety of Vehicles},
  \bibinfo{organization}{National Highway Traffic Safety Administration
  Washington, DC}.
\bibitem[{Sun and Kondyli(2010)}]{sun2010modeling}
\bibinfo{author}{Sun, D.}, \bibinfo{author}{Kondyli, A.}, \bibinfo{year}{2010}.
\newblock \bibinfo{title}{Modeling vehicle interactions during lane-changing
  behavior on arterial streets}.
\newblock \bibinfo{journal}{Computer-Aided Civil and Infrastructure
  Engineering} \bibinfo{volume}{25}, \bibinfo{pages}{557--571}.
\bibitem[{Sv{\"a}rd et~al.(2021)Sv{\"a}rd, Markkula, B{\"a}rgman and
  Victor}]{svard2021computational}
\bibinfo{author}{Sv{\"a}rd, M.}, \bibinfo{author}{Markkula, G.},
  \bibinfo{author}{B{\"a}rgman, J.}, \bibinfo{author}{Victor, T.},
  \bibinfo{year}{2021}.
\newblock \bibinfo{title}{Computational modeling of driver pre-crash brake
  response, with and without off-road glances: Parameterization using
  real-world crashes and near-crashes}.
\newblock \bibinfo{journal}{Accident Analysis \& Prevention}
  \bibinfo{volume}{163}, \bibinfo{pages}{106433}.
\bibitem[{Sv{\"a}rd et~al.(2017)Sv{\"a}rd, Markkula, Engstr{\"o}m, Granum and
  B{\"a}rgman}]{svard2017quantitative}
\bibinfo{author}{Sv{\"a}rd, M.}, \bibinfo{author}{Markkula, G.},
  \bibinfo{author}{Engstr{\"o}m, J.}, \bibinfo{author}{Granum, F.},
  \bibinfo{author}{B{\"a}rgman, J.}, \bibinfo{year}{2017}.
\newblock \bibinfo{title}{A quantitative driver model of pre-crash brake onset
  and control}, in: \bibinfo{booktitle}{Proceedings of the Human Factors and
  Ergonomics Society Annual Meeting}, \bibinfo{organization}{SAGE Publications
  Sage CA: Los Angeles, CA}. pp. \bibinfo{pages}{339--343}.
\bibitem[{Tian et~al.(2021)Tian, Cao, Huang, Fei, Zheng and
  Ji}]{tian2021learning}
\bibinfo{author}{Tian, Y.}, \bibinfo{author}{Cao, X.}, \bibinfo{author}{Huang,
  K.}, \bibinfo{author}{Fei, C.}, \bibinfo{author}{Zheng, Z.},
  \bibinfo{author}{Ji, X.}, \bibinfo{year}{2021}.
\newblock \bibinfo{title}{Learning to drive like human beings: A method based
  on deep reinforcement learning}.
\newblock \bibinfo{journal}{IEEE Transactions on Intelligent Transportation
  Systems} .
\bibitem[{Tseng et~al.(2005)Tseng, Asgari, Hrovat, Van Der~Jagt, Cherry and
  Neads}]{tseng2005evasive}
\bibinfo{author}{Tseng, H.}, \bibinfo{author}{Asgari, J.},
  \bibinfo{author}{Hrovat, D.}, \bibinfo{author}{Van Der~Jagt, P.},
  \bibinfo{author}{Cherry, A.}, \bibinfo{author}{Neads, S.},
  \bibinfo{year}{2005}.
\newblock \bibinfo{title}{Evasive manoeuvres with a steering robot}.
\newblock \bibinfo{journal}{Vehicle system dynamics} \bibinfo{volume}{43},
  \bibinfo{pages}{199--216}.
\bibitem[{Uhlenbeck and Ornstein(1930)}]{uhlenbeck1930theory}
\bibinfo{author}{Uhlenbeck, G.E.}, \bibinfo{author}{Ornstein, L.S.},
  \bibinfo{year}{1930}.
\newblock \bibinfo{title}{On the theory of the brownian motion}.
\newblock \bibinfo{journal}{Physical review} \bibinfo{volume}{36},
  \bibinfo{pages}{823}.
\bibitem[{Van Der~Horst and Hogema(1993)}]{van1993time}
\bibinfo{author}{Van Der~Horst, R.}, \bibinfo{author}{Hogema, J.},
  \bibinfo{year}{1993}.
\newblock \bibinfo{title}{Time-to-collision and collision avoidance systems} .
\bibitem[{Van~Rossum and Drake(2009)}]{python}
\bibinfo{author}{Van~Rossum, G.}, \bibinfo{author}{Drake, F.L.},
  \bibinfo{year}{2009}.
\newblock \bibinfo{title}{Python 3 Reference Manual}.
\newblock \bibinfo{publisher}{CreateSpace}, \bibinfo{address}{Scotts Valley,
  CA}.
\bibitem[{Venkatraman et~al.(2016)Venkatraman, Lee and
  Schwarz}]{venkatraman2016steer}
\bibinfo{author}{Venkatraman, V.}, \bibinfo{author}{Lee, J.D.},
  \bibinfo{author}{Schwarz, C.W.}, \bibinfo{year}{2016}.
\newblock \bibinfo{title}{Steer or brake?: Modeling drivers’
  collision-avoidance behavior by using perceptual cues}.
\newblock \bibinfo{journal}{Transportation research record}
  \bibinfo{volume}{2602}, \bibinfo{pages}{97--103}.
\bibitem[{Venthuruthiyil and Chunchu(2022)}]{venthuruthiyil2022anticipated}
\bibinfo{author}{Venthuruthiyil, S.P.}, \bibinfo{author}{Chunchu, M.},
  \bibinfo{year}{2022}.
\newblock \bibinfo{title}{Anticipated collision time (act): A two-dimensional
  surrogate safety indicator for trajectory-based proactive safety assessment}.
\newblock \bibinfo{journal}{Transportation Research Part C: Emerging
  Technologies} \bibinfo{volume}{139}, \bibinfo{pages}{103655}.
\bibitem[{Wang and Coifman(2008)}]{wang2008effect}
\bibinfo{author}{Wang, C.}, \bibinfo{author}{Coifman, B.},
  \bibinfo{year}{2008}.
\newblock \bibinfo{title}{The effect of lane-change maneuvers on a simplified
  car-following theory}.
\newblock \bibinfo{journal}{IEEE transactions on intelligent transportation
  systems} \bibinfo{volume}{9}, \bibinfo{pages}{523--535}.
\bibitem[{Wang et~al.(2018)Wang, Xu, Xia, Qian and Lu}]{wang2018combined}
\bibinfo{author}{Wang, C.}, \bibinfo{author}{Xu, C.}, \bibinfo{author}{Xia,
  J.}, \bibinfo{author}{Qian, Z.}, \bibinfo{author}{Lu, L.},
  \bibinfo{year}{2018}.
\newblock \bibinfo{title}{A combined use of microscopic traffic simulation and
  extreme value methods for traffic safety evaluation}.
\newblock \bibinfo{journal}{Transportation Research Part C: Emerging
  Technologies} \bibinfo{volume}{90}, \bibinfo{pages}{281--291}.
\bibitem[{Wang et~al.(2021)Wang, Hu, Li and Li}]{wang2021harmonious}
\bibinfo{author}{Wang, G.}, \bibinfo{author}{Hu, J.}, \bibinfo{author}{Li, Z.},
  \bibinfo{author}{Li, L.}, \bibinfo{year}{2021}.
\newblock \bibinfo{title}{Harmonious lane changing via deep reinforcement
  learning}.
\newblock \bibinfo{journal}{IEEE Transactions on Intelligent Transportation
  Systems} .
\bibitem[{Ward et~al.(2015)Ward, Agamennoni, Worrall, Bender and
  Nebot}]{ward2015extending}
\bibinfo{author}{Ward, J.R.}, \bibinfo{author}{Agamennoni, G.},
  \bibinfo{author}{Worrall, S.}, \bibinfo{author}{Bender, A.},
  \bibinfo{author}{Nebot, E.}, \bibinfo{year}{2015}.
\newblock \bibinfo{title}{Extending time to collision for probabilistic
  reasoning in general traffic scenarios}.
\newblock \bibinfo{journal}{Transportation Research Part C: Emerging
  Technologies} \bibinfo{volume}{51}, \bibinfo{pages}{66--82}.
\bibitem[{Xie et~al.(2016)Xie, Li, Ozbay, Dobler, Yang, Chiang and
  Ghandehari}]{xie2016development}
\bibinfo{author}{Xie, K.}, \bibinfo{author}{Li, C.}, \bibinfo{author}{Ozbay,
  K.}, \bibinfo{author}{Dobler, G.}, \bibinfo{author}{Yang, H.},
  \bibinfo{author}{Chiang, A.T.}, \bibinfo{author}{Ghandehari, M.},
  \bibinfo{year}{2016}.
\newblock \bibinfo{title}{Development of a comprehensive framework for
  video-based safety assessment}, in: \bibinfo{booktitle}{2016 IEEE 19th
  International Conference on Intelligent Transportation Systems (ITSC)},
  \bibinfo{organization}{IEEE}. pp. \bibinfo{pages}{2638--2643}.
\bibitem[{Xie et~al.(2019)Xie, Yang, Ozbay and Yang}]{xie2019use}
\bibinfo{author}{Xie, K.}, \bibinfo{author}{Yang, D.}, \bibinfo{author}{Ozbay,
  K.}, \bibinfo{author}{Yang, H.}, \bibinfo{year}{2019}.
\newblock \bibinfo{title}{Use of real-world connected vehicle data in
  identifying high-risk locations based on a new surrogate safety measure}.
\newblock \bibinfo{journal}{Accident Analysis \& Prevention}
  \bibinfo{volume}{125}, \bibinfo{pages}{311--319}.
\bibitem[{Xing et~al.(2019)Xing, He, Abdel-Aty, Cai, Li and
  Zheng}]{xing2019examining}
\bibinfo{author}{Xing, L.}, \bibinfo{author}{He, J.},
  \bibinfo{author}{Abdel-Aty, M.}, \bibinfo{author}{Cai, Q.},
  \bibinfo{author}{Li, Y.}, \bibinfo{author}{Zheng, O.}, \bibinfo{year}{2019}.
\newblock \bibinfo{title}{Examining traffic conflicts of up stream toll plaza
  area using vehicles’ trajectory data}.
\newblock \bibinfo{journal}{Accident Analysis \& Prevention}
  \bibinfo{volume}{125}, \bibinfo{pages}{174--187}.
\bibitem[{Xing et~al.(2020)Xing, Lv, Wang, Cao and Velenis}]{xing2020ensemble}
\bibinfo{author}{Xing, Y.}, \bibinfo{author}{Lv, C.}, \bibinfo{author}{Wang,
  H.}, \bibinfo{author}{Cao, D.}, \bibinfo{author}{Velenis, E.},
  \bibinfo{year}{2020}.
\newblock \bibinfo{title}{An ensemble deep learning approach for driver lane
  change intention inference}.
\newblock \bibinfo{journal}{Transportation Research Part C: Emerging
  Technologies} \bibinfo{volume}{115}, \bibinfo{pages}{102615}.
\bibitem[{Xiong et~al.(2019)Xiong, Wang, Cai, Chen, Farah and
  Hagenzieker}]{xiong2019forward}
\bibinfo{author}{Xiong, X.}, \bibinfo{author}{Wang, M.}, \bibinfo{author}{Cai,
  Y.}, \bibinfo{author}{Chen, L.}, \bibinfo{author}{Farah, H.},
  \bibinfo{author}{Hagenzieker, M.}, \bibinfo{year}{2019}.
\newblock \bibinfo{title}{A forward collision avoidance algorithm based on
  driver braking behavior}.
\newblock \bibinfo{journal}{Accident Analysis \& Prevention}
  \bibinfo{volume}{129}, \bibinfo{pages}{30--43}.
\bibitem[{Xu et~al.(2012)Xu, Liu, Ou and Song}]{xu2012dynamic}
\bibinfo{author}{Xu, G.}, \bibinfo{author}{Liu, L.}, \bibinfo{author}{Ou, Y.},
  \bibinfo{author}{Song, Z.}, \bibinfo{year}{2012}.
\newblock \bibinfo{title}{Dynamic modeling of driver control strategy of
  lane-change behavior and trajectory planning for collision prediction}.
\newblock \bibinfo{journal}{IEEE Transactions on Intelligent Transportation
  Systems} \bibinfo{volume}{13}, \bibinfo{pages}{1138--1155}.
\bibitem[{Xue et~al.(2018)Xue, Markkula, Yan and Merat}]{xue2018using}
\bibinfo{author}{Xue, Q.}, \bibinfo{author}{Markkula, G.},
  \bibinfo{author}{Yan, X.}, \bibinfo{author}{Merat, N.}, \bibinfo{year}{2018}.
\newblock \bibinfo{title}{Using perceptual cues for brake response to a lead
  vehicle: Comparing threshold and accumulator models of visual looming}.
\newblock \bibinfo{journal}{Accident Analysis \& Prevention}
  \bibinfo{volume}{118}, \bibinfo{pages}{114--124}.
\bibitem[{Yang et~al.(2021a)Yang, Ozbay, Xie, Yang, Zuo and
  Sha}]{yang2021proactive}
\bibinfo{author}{Yang, D.}, \bibinfo{author}{Ozbay, K.}, \bibinfo{author}{Xie,
  K.}, \bibinfo{author}{Yang, H.}, \bibinfo{author}{Zuo, F.},
  \bibinfo{author}{Sha, D.}, \bibinfo{year}{2021}a.
\newblock \bibinfo{title}{Proactive safety monitoring: A functional approach to
  detect safety-related anomalies using unmanned aerial vehicle video data}.
\newblock \bibinfo{journal}{Transportation research part C: emerging
  technologies} \bibinfo{volume}{127}, \bibinfo{pages}{103130}.
\bibitem[{Yang et~al.(2021b)Yang, Xie, Ozbay and Yang}]{yang2021fusing}
\bibinfo{author}{Yang, D.}, \bibinfo{author}{Xie, K.}, \bibinfo{author}{Ozbay,
  K.}, \bibinfo{author}{Yang, H.}, \bibinfo{year}{2021}b.
\newblock \bibinfo{title}{Fusing crash data and surrogate safety measures for
  safety assessment: Development of a structural equation model with
  conditional autoregressive spatial effect and random parameters}.
\newblock \bibinfo{journal}{Accident Analysis \& Prevention}
  \bibinfo{volume}{152}, \bibinfo{pages}{105971}.
\bibitem[{Ye et~al.(2019)Ye, Zhang and Sun}]{ye2019automated}
\bibinfo{author}{Ye, Y.}, \bibinfo{author}{Zhang, X.}, \bibinfo{author}{Sun,
  J.}, \bibinfo{year}{2019}.
\newblock \bibinfo{title}{Automated vehicle’s behavior decision making using
  deep reinforcement learning and high-fidelity simulation environment}.
\newblock \bibinfo{journal}{Transportation Research Part C: Emerging
  Technologies} \bibinfo{volume}{107}, \bibinfo{pages}{155--170}.
\bibitem[{Yuan et~al.(2019)Yuan, Sun and Gordon}]{yuan2019unified}
\bibinfo{author}{Yuan, H.}, \bibinfo{author}{Sun, X.}, \bibinfo{author}{Gordon,
  T.}, \bibinfo{year}{2019}.
\newblock \bibinfo{title}{Unified decision-making and control for highway
  collision avoidance using active front steer and individual wheel torque
  control}.
\newblock \bibinfo{journal}{Vehicle system dynamics} \bibinfo{volume}{57},
  \bibinfo{pages}{1188--1205}.
\bibitem[{Zaharia et~al.(2016)Zaharia, Xin, Wendell, Das, Armbrust, Dave, Meng,
  Rosen, Venkataraman, Franklin et~al.}]{zaharia2016apache}
\bibinfo{author}{Zaharia, M.}, \bibinfo{author}{Xin, R.S.},
  \bibinfo{author}{Wendell, P.}, \bibinfo{author}{Das, T.},
  \bibinfo{author}{Armbrust, M.}, \bibinfo{author}{Dave, A.},
  \bibinfo{author}{Meng, X.}, \bibinfo{author}{Rosen, J.},
  \bibinfo{author}{Venkataraman, S.}, \bibinfo{author}{Franklin, M.J.}, et~al.,
  \bibinfo{year}{2016}.
\newblock \bibinfo{title}{Apache spark: a unified engine for big data
  processing}.
\newblock \bibinfo{journal}{Communications of the ACM} \bibinfo{volume}{59},
  \bibinfo{pages}{56--65}.
\bibitem[{Zhang et~al.(2021)Zhang, Zhu, Wang and Xi}]{zhang2021spatiotemporal}
\bibinfo{author}{Zhang, C.}, \bibinfo{author}{Zhu, J.}, \bibinfo{author}{Wang,
  W.}, \bibinfo{author}{Xi, J.}, \bibinfo{year}{2021}.
\newblock \bibinfo{title}{Spatiotemporal learning of multivehicle interaction
  patterns in lane-change scenarios}.
\newblock \bibinfo{journal}{IEEE Transactions on Intelligent Transportation
  Systems} .
\bibitem[{Zhao et~al.(2017)Zhao, Guo and Jia}]{zhao2017trafficnet}
\bibinfo{author}{Zhao, D.}, \bibinfo{author}{Guo, Y.}, \bibinfo{author}{Jia,
  Y.J.}, \bibinfo{year}{2017}.
\newblock \bibinfo{title}{Trafficnet: An open naturalistic driving scenario
  library}, in: \bibinfo{booktitle}{2017 IEEE 20th International Conference on
  Intelligent Transportation Systems (ITSC)}, \bibinfo{organization}{IEEE}. pp.
  \bibinfo{pages}{1--8}.
\bibitem[{Zheng et~al.(2020)Zheng, Zeng, Yang, Li and Zhan}]{zheng2020bezier}
\bibinfo{author}{Zheng, L.}, \bibinfo{author}{Zeng, P.}, \bibinfo{author}{Yang,
  W.}, \bibinfo{author}{Li, Y.}, \bibinfo{author}{Zhan, Z.},
  \bibinfo{year}{2020}.
\newblock \bibinfo{title}{B{\'e}zier curve-based trajectory planning for
  autonomous vehicles with collision avoidance}.
\newblock \bibinfo{journal}{IET Intelligent Transport Systems}
  \bibinfo{volume}{14}, \bibinfo{pages}{1882--1891}.
\bibitem[{Zheng(2014)}]{zheng2014recent}
\bibinfo{author}{Zheng, Z.}, \bibinfo{year}{2014}.
\newblock \bibinfo{title}{Recent developments and research needs in modeling
  lane changing}.
\newblock \bibinfo{journal}{Transportation research part B: methodological}
  \bibinfo{volume}{60}, \bibinfo{pages}{16--32}.
\bibitem[{Zheng et~al.(2013)Zheng, Ahn, Chen and Laval}]{zheng2013effects}
\bibinfo{author}{Zheng, Z.}, \bibinfo{author}{Ahn, S.}, \bibinfo{author}{Chen,
  D.}, \bibinfo{author}{Laval, J.}, \bibinfo{year}{2013}.
\newblock \bibinfo{title}{The effects of lane-changing on the immediate
  follower: Anticipation, relaxation, and change in driver characteristics}.
\newblock \bibinfo{journal}{Transportation research part C: emerging
  technologies} \bibinfo{volume}{26}, \bibinfo{pages}{367--379}.
\bibitem[{Zheng et~al.(2010)Zheng, Ahn and Monsere}]{zheng2010impact}
\bibinfo{author}{Zheng, Z.}, \bibinfo{author}{Ahn, S.},
  \bibinfo{author}{Monsere, C.M.}, \bibinfo{year}{2010}.
\newblock \bibinfo{title}{Impact of traffic oscillations on freeway crash
  occurrences}.
\newblock \bibinfo{journal}{Accident Analysis \& Prevention}
  \bibinfo{volume}{42}, \bibinfo{pages}{626--636}.
\bibitem[{Zhou and Zhong(2020)}]{zhou2020evasive}
\bibinfo{author}{Zhou, H.}, \bibinfo{author}{Zhong, Z.}, \bibinfo{year}{2020}.
\newblock \bibinfo{title}{Evasive behavior-based method for threat assessment
  in different scenarios: A novel framework for intelligent vehicle}.
\newblock \bibinfo{journal}{Accident Analysis \& Prevention}
  \bibinfo{volume}{148}, \bibinfo{pages}{105798}.
\bibitem[{Zhu et~al.(2018)Zhu, Wang and Wang}]{zhu2018human}
\bibinfo{author}{Zhu, M.}, \bibinfo{author}{Wang, X.}, \bibinfo{author}{Wang,
  Y.}, \bibinfo{year}{2018}.
\newblock \bibinfo{title}{Human-like autonomous car-following model with deep
  reinforcement learning}.
\newblock \bibinfo{journal}{Transportation research part C: emerging
  technologies} \bibinfo{volume}{97}, \bibinfo{pages}{348--368}.
\bibitem[{Zhu et~al.(2020)Zhu, Wang, Pu, Hu, Wang and Ke}]{zhu2020safe}
\bibinfo{author}{Zhu, M.}, \bibinfo{author}{Wang, Y.}, \bibinfo{author}{Pu,
  Z.}, \bibinfo{author}{Hu, J.}, \bibinfo{author}{Wang, X.},
  \bibinfo{author}{Ke, R.}, \bibinfo{year}{2020}.
\newblock \bibinfo{title}{Safe, efficient, and comfortable velocity control
  based on reinforcement learning for autonomous driving}.
\newblock \bibinfo{journal}{Transportation Research Part C: Emerging
  Technologies} \bibinfo{volume}{117}, \bibinfo{pages}{102662}.
\bibitem[{Zuo et~al.(2020)Zuo, Ozbay, Kurkcu, Gao, Yang and
  Xie}]{zuo2020microscopic}
\bibinfo{author}{Zuo, F.}, \bibinfo{author}{Ozbay, K.},
  \bibinfo{author}{Kurkcu, A.}, \bibinfo{author}{Gao, J.},
  \bibinfo{author}{Yang, H.}, \bibinfo{author}{Xie, K.}, \bibinfo{year}{2020}.
\newblock \bibinfo{title}{Microscopic simulation based study of pedestrian
  safety applications at signalized urban crossings in a connected-automated
  vehicle environment and reinforcement learning based optimization of vehicle
  decisions.}
\newblock \bibinfo{journal}{Advances in transportation studies} .

\end{thebibliography}

\end{document}